\documentclass{article} 
\usepackage[final]{colm2026_conference}
\usepackage[most]{tcolorbox}
\usepackage[table]{xcolor}
\newcommand{\masktok}[1]{\colorbox{gray!20}{\ttfamily[#1]}}
\usepackage{tabularx}
\usepackage{booktabs}
\usepackage{microtype}
\usepackage{hyperref}
\usepackage{url}
\usepackage{amsmath,amssymb}
\usepackage{graphicx}
\usepackage{wrapfig}
\usepackage{capt-of}
\usepackage{algorithm}
\usepackage{algpseudocode}
\usepackage{array}
\usepackage{placeins}
\newcolumntype{C}{>{\centering\arraybackslash}X}

\definecolor{darkblue}{rgb}{0, 0, 0.5}
\hypersetup{colorlinks=true, citecolor=darkblue, linkcolor=darkblue, urlcolor=darkblue}

\usepackage{colortbl}

\title{MetaState: Persistent Working Memory Enhances Reasoning in Discrete Diffusion Language Models}

\author{{\bf Kejing Xia$^{1}$,} {\bf Mingzhe Li$^{2}$,} {\bf Lixuan Wei$^{3}$,} {\bf Zhenbang Du$^{1}$,} {\bf Xiangchi Yuan$^{1}$,} \\ {\bf Dachuan Shi$^{1}$,} {\bf Qirui Jin$^{1}$,} {\bf Wenke Lee$^{1}$} \\
$^{1}$Georgia Institute of Technology  $^{2}$University of Massachusetts Amherst \\ $^{3}$Harvard University}

\begin{document}

\maketitle

\begin{abstract}
Discrete diffusion language models (dLLMs) generate text by iteratively denoising a masked sequence.  However, standard dLLMs condition each denoising step solely on the current hard-masked sequence, while intermediate continuous representations are discarded after sampling and remasking. We term this bottleneck the \textbf{Information Island} issue: continuous information remains isolated within individual denoising steps and fails to propagate across the trajectory. This bottleneck is especially harmful for reasoning, which requires intermediate reasoning state to be preserved and updated across many denoising steps. To address this limitation, we introduce \textbf{MetaState}, a lightweight recurrent augmentation that equips a frozen dLLM backbone with persistent, fixed-size working memory. MetaState comprises three modules with a shared time conditioner: a cross-attention \textbf{Mixer} that reads backbone activations into memory slots, a GRU-style \textbf{Updater} that integrates information across steps, and a cross-attention \textbf{Injector} that writes the updated memory back into the backbone. We train these modules with a dedicated $K$-step unrolling pipeline to learn multi-step dynamics. MetaState adds only ${\sim}0.6\%$ trainable parameters while keeping the backbone frozen, and consistently improves reasoning performance over frozen baselines on mathematical reasoning and code generation benchmarks, with an average gain of 4.5 percentage points across all evaluations.
Our code is available at \url{https://github.com/Les1a/MetaState}.

\end{abstract}

\section{Introduction}
\label{sec:intro}
\begin{wrapfigure}{r}{0.473\textwidth}
    \centering
    \vspace{-\intextsep}
    \includegraphics[width=\linewidth]{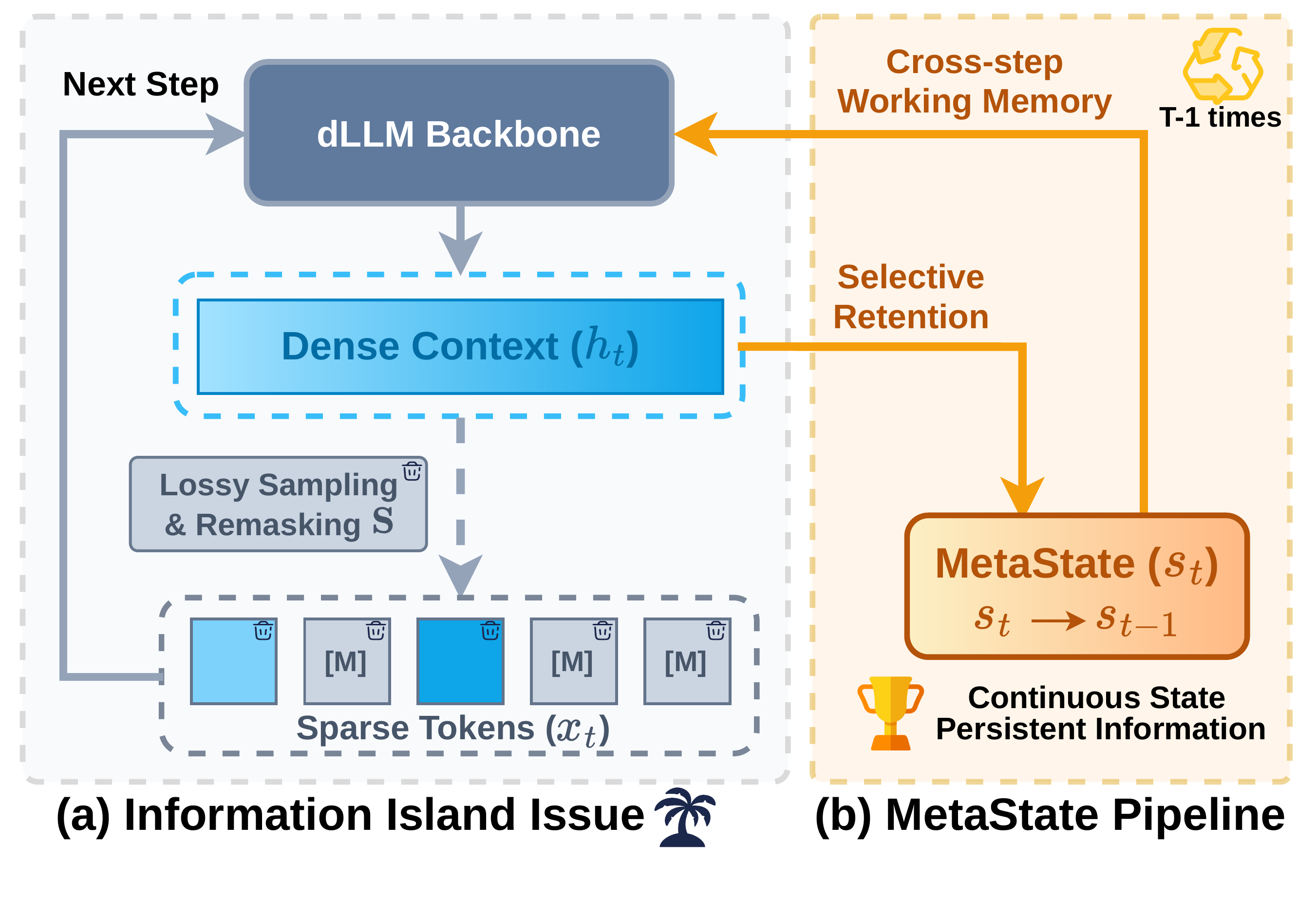}
    \vspace{-2.8em}
    \caption{The Information Island issue in discrete diffusion: sampling and remasking compress continuous hidden activations into discrete tokens, imposing a lossy bottleneck between denoising steps. MetaState addresses this issue by maintaining a persistent state across steps.}
    \label{fig:teaser1}
    \vspace{-\intextsep}
\end{wrapfigure}

Autoregressive (AR) language models factorize the joint distribution over sequences into a product of conditional probabilities, producing one token per forward pass~\citep{radford2018improving, radford2019language, brown2020language}. Although this paradigm underlies many recent foundation models, the left-to-right causal structure prevents parallel decoding and limits the use of bidirectional context. Discrete diffusion language models (dLLMs) have recently emerged as a non-autoregressive alternative~\citep{li2025survey, sahoo2024simple, gong2024scaling}. Starting from a fully corrupted sequence, dLLMs iteratively denoise to recover clean text and update arbitrary positions using bidirectional attention~\citep{austin2021structured}. When scaled to billions of parameters, dLLMs achieve quality comparable to that of autoregressive models while retaining the advantages of decoding parallelism and generation flexibility, as demonstrated by the LLaDA series~\citep{nie2025large,zhu2025llada,bie2025llada2} and Dream~\citep{ye2025dream}.

\begin{figure}[t]
    \centering
    \includegraphics[width=1\linewidth]{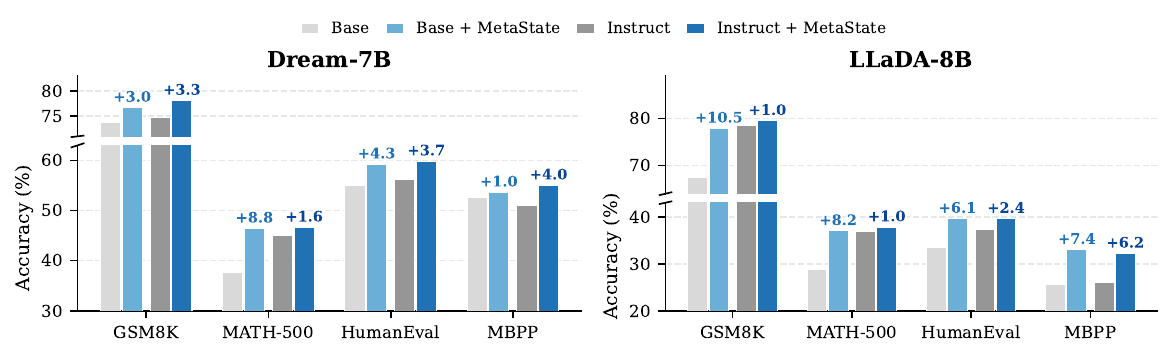}
    \vspace{-2.4em}
    \caption{Performance comparison between MetaState and frozen baselines on reasoning benchmarks for LLaDA-8B and Dream-7B in both Instruct and Base versions.}
    \label{fig:teaser2}
    \vspace{-1.0em}
\end{figure}

Current standard dLLMs nevertheless share a limitation that we term the Information Island issue. In the diffusion formulation, the inter-step process is Markovian in the discrete sequence state: each transition conditions on the current masked tokens $\mathbf{x}_t$, while the continuous hidden representation $\mathbf{h}_t$ computed at that step is not carried forward explicitly,
\vspace{-0.4em}
\begin{equation}
\label{eq:information_island}
\begin{aligned}
p_\theta(\mathbf{x}_{0:T}) = p(\mathbf{x}_T)\prod_{t=1}^{T} p_\theta(\mathbf{x}_{t-1} \mid \mathbf{x}_t), \quad
\mathbf{x}_{t-1} = \mathcal{S}\!\left( \mathbf{h}_t\right).
\end{aligned}
\end{equation}
At each denoising step $t$, the model computes a high-dimensional hidden representation $\mathbf{h}_t$ that encodes substantially richer information than the discrete tokens passed to the next step. Beyond token-level predictive semantics, $\mathbf{h}_t$ also captures long-range dependencies and global sequence structure information. However, the sampling-and-remasking interface $\mathcal{S}$ maps this rich continuous state to discrete token identities and sparse remasking indicators. This transition discards the continuous information in $\mathbf{h}_t$ and compresses each step's computation into a sparse discrete sequence. As a result, the next denoising step receives only a highly lossy representation of the information computed at the previous step. We refer to this repeated cross-step information loss as the Information Island issue.

This bottleneck arises at every transition along the denoising trajectory, as illustrated in Fig.~\ref{fig:teaser1}. In the diffusion process, early high-noise steps often establish coarse global structure, while later low-noise steps refine local details and enforce fine-grained constraints. However, useful inferences made at one step must be reconstructed from the sparse token sequence at later steps. This repeated reconstruction can introduce cross-step drift: intermediate information may be weakened, overwritten, or inconsistently re-derived as denoising proceeds. Such drift is especially harmful for reasoning, where success depends on preserving intermediate computations in multi-step mathematical reasoning and maintaining global program constraints such as variable scope and control flow across long denoising trajectories.

To address this limitation, we propose MetaState, a lightweight recurrent augmentation that equips a frozen dLLM backbone with persistent working memory across denoising steps. Motivated by evidence that working memory capacity is an important factor in language model reasoning~\citep{zhang2024working}, MetaState maintains a compact set of continuous memory slots that persist across denoising steps, thereby adding a parallel information path alongside the standard discrete denoising path. Concretely, three lightweight modules form a recurrent loop around the frozen backbone: a Mixer reads backbone activations into the memory slots, an Updater integrates newly extracted information through gated recurrence, and an Injector writes the updated state back into the backbone's input embeddings for the next step. A shared time conditioner coordinates all three modules. To train this recurrent memory to retain and update information across steps, we further introduce a dedicated $K$-step unrolling procedure that backpropagates through the denoising trajectory (\S\ref{sec:training}).
In summary, our contributions are as follows:
\begin{enumerate}
    \item We identify the Information Island issue in discrete diffusion language models, a representational bottleneck in which rich hidden activations are compressed into sparse and discrete tokens at every denoising step, and analyze why this issue is especially harmful for multi-step reasoning.
    \item We propose \textbf{MetaState}, a backbone-agnostic recurrent augmentation that maintains constant-size persistent working memory throughout the denoising process, along with a $K$-step unrolling training procedure that enables gradient flow through the multi-step state trajectory.
    \item We validate MetaState on two distinct dLLM backbones, LLaDA-8B~\citep{nie2025large} and Dream-7B~\citep{ye2025dream}, over standard mathematical reasoning and code generation benchmarks (GSM8K, MATH-500, HumanEval, and MBPP), achieving an average improvement of 4.5 percentage points at negligible parameter cost.
\end{enumerate}
\section{Related Work}
\label{sec:related}

\subsection{Discrete Diffusion LLMs}
Discrete diffusion LLMs formulate text generation as a non-autoregressive process by adapting diffusion dynamics to discrete token spaces. D3PM~\citep{austin2021structured} formalized this approach using discrete transition matrices. Subsequent models largely adopt the masked diffusion paradigm: MDLM~\citep{sahoo2024simple} derives a variational lower bound and SEDD~\citep{lou2023discrete} introduces a score entropy objective, enabling masked dLLMs to reach likelihood and perplexity levels comparable to those of autoregressive (AR) models. At the billion-parameter scale, LLaDA series~\citep{nie2025large,zhu2025llada,zhu2025lladamoe} and Dream~\citep{ye2025dream} demonstrate that dLLMs can match AR quality while enabling parallel decoding and bidirectional attention. Semi-AR methods like BD3-LMs~\citep{arriola2025block} and SDAR~\citep{cheng2025sdar} combine inter-block autoregression with intra-block parallel diffusion. For cache integration, dLLM-Cache~\citep{liu2025dllm} targets dual redundancy in prompts and responses via adaptive caching, while Fast-dLLM~\citep{wu2025fast} leverages activation similarity for block-wise KV-cache reuse.

\subsection{Continuous Diffusion and Latent Reasoning}

To fully use continuous context, prior work proposes modifying either the diffusion kernel or the decoding interface. CADD and CANDI~\citep{zheng2025continuously, pynadath2025candi} couple discrete masking with continuous diffusion through hybrid kernels, changing the forward and reverse diffusion formulation and training the backbone to operate under that process. MetaState targets a different level: it keeps the original discrete diffusion backbone frozen and adds an external recurrent state at the sampling-and-remasking boundary. Other decoding-side methods such as LRD and RCD~\citep{zhu2025latent, hu2026residual} replace hard token sampling with probability mixtures, whereas MetaState leaves the discrete path intact and adds a parallel cross-step memory path. DCoLT~\citep{huang2025reinforcing} designs a trajectory-level reasoning policy.
In continuous image diffusion, Recurrent Interface Networks~\citep{jabri2022scalable} maintain persistent latent tokens across denoising steps, and Diffusion Forcing~\citep{chen2024diffusion} couples an RNN with the diffusion process.
In AR modeling, several latent reasoning methods propagate continuous latent representations to bypass discrete decoding bottlenecks, including Coconut, CODI, Soft Thinking and SwiReasoning~\citep{hao2024training, shen2025codi, zhang2025soft, shi2025swireasoning}.
Building on this line of work, LaDiR and STAR-LDM~\citep{kang2025ladir, lovelace2026stop} use latent diffusion for trajectory planning. However, these techniques only apply to sequential generation and do not transfer to the dLLM paradigm. As a result, dLLMs still lack a frozen-backbone mechanism that maintains continuous memory across diffusion steps to address the Information Island issue.
\section{Preliminaries}
\label{sec:preliminaries}

We consider the masked diffusion paradigm~\citep{austin2021structured, ou2024your}, which defines a forward masking process that progressively replaces tokens with a special $[\text{MASK}]$ token $\mathbf{M}$ and a reverse unmasking process that recovers the original sequence.

\textbf{Forward process.} Given a clean sequence $\mathbf{x}_0 = (x_0^{(1)}, \ldots, x_0^{(N)})$ over vocabulary $\mathcal{V}$, the forward process produces a noisy sequence $\mathbf{x}_t$ by independently masking each token with probability $1 - \alpha_t$, where $\alpha_t$ denotes the token retention probability from the discrete diffusion convention~\citep{shi2024simplified, nie2025large}. In continuous time $t \in (0, 1]$, the schedule decreases from $\alpha_0 = 1$ (fully clean) to $\alpha_1 \approx 0$ (fully masked). Each token is kept unchanged with probability \(\alpha_t\), and is replaced by the mask token \(\mathbf{M}\) with probability \(1-\alpha_t\).

\textbf{Reverse process.} For any noise level $t \in (0, 1]$, the model $p_\theta$ predicts the clean token at each masked position $i$ where $x_t^{(i)} = \mathbf{M}$, yielding the conditional distribution $p_\theta(x_0^{(i)} \mid \mathbf{x}_t)$.

\textbf{Training objective.} The training loss is the expected cross-entropy over masked positions:
\begin{align*}
\mathcal{L}_\text{MDLM} = \mathbb{E}_{t \sim \mathcal{U}(0, 1],\, \mathbf{x}_0,\, \mathbf{x}_t \sim q(\mathbf{x}_t \mid \mathbf{x}_0)} \!\left[ \frac{1}{t}\!\sum_{i:\, x_t^{(i)} = \mathbf{M}}\!- \log p_\theta(x_0^{(i)} \mid \mathbf{x}_t) \right].
\end{align*}

\textbf{Inference.} Starting from a fully masked sequence, the time is discretized into $T$ steps. At each discrete step $t$, the model samples clean tokens and selectively remasks a subset of positions according to prediction confidence, yielding a progressively cleaner sequence $\mathbf{x}_{t-1}$.

\textbf{Information Island issue.} As discussed in \S\ref{sec:intro}, this formulation gives rise to the Information Island issue: the sampling and remasking operator discards continuous hidden activations at each step (see Appendix~\ref{sec:info-island} for a detailed analysis).

\section{Method}
\label{sec:method}

\begin{figure}[t]
    \centering
    \includegraphics[width=1\linewidth]{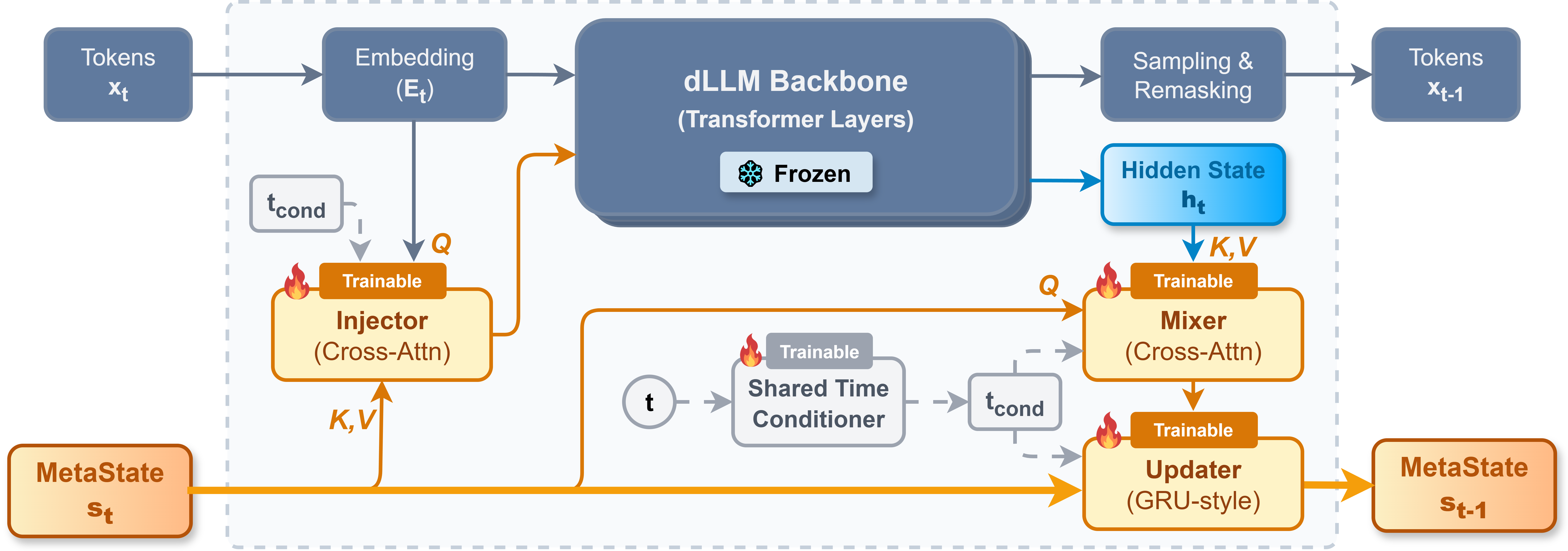}
    \vspace{-2.0em}
    \caption{Overview of the MetaState architecture. The three modules (Injector, Mixer, Updater) and the shared time conditioner form a recurrent loop around the frozen backbone, propagating a persistent state across denoising steps.}
    \label{fig:method}
    \vspace{-1.5em}
\end{figure}

\subsection{MetaState Overview}
\label{sec:metastate-overview}

To resolve the Information Island issue, MetaState introduces a continuous memory that persists across the discrete interface steps. This memory is maintained by three lightweight modules, the \emph{Injector}, \emph{Mixer}, and \emph{Updater}, coordinated by a shared time conditioner (Figure~\ref{fig:method}). All modules operate in bottleneck dimensions with a fixed slot count independent of sequence length. The complete pipeline is given in Algorithm~\ref{alg:metastate-step} (Appendix).

Because the discrete interface between successive denoising steps discards all intermediate representations, each step has no access to the processing history accumulated by prior steps. To bridge this gap, we augment the denoising process with a persistent state $\mathbf{s}_t \in \mathbb{R}^{M \times D_s}$, organized as $M$ fixed memory slots of dimension $D_s$. This fixed-size design is critical: it ensures that the memory overhead does not grow with the number of tokens, and it encourages the network to learn a compact representation of the generation trajectory rather than simply caching raw activations. The augmented transition becomes:
\begin{align*}
p_\theta(\mathbf{x}_{t-1}\mid \mathbf{x}_t,\mathbf{s}_t),
\qquad
\mathbf{s}_{t-1} = g_\theta(\mathbf{s}_t,\mathbf{h}_t,t),
\end{align*}
where $g_\theta$ denotes the state-update function realized by the Mixer and Updater. The state requires $\mathcal{O}(M D_s)$ storage and does not scale with the sequence length $N$. In addition, each cross-attention operation is performed with a fixed number of memory slots in a bottleneck dimension. Thus the overall overhead is dominated by the cost of the frozen backbone.

At each denoising step, the Injector first writes the current state (when the state exists) into the backbone's input embeddings using the conditioning feature from the end of the previous step. After the backbone forward pass, the Mixer reads the final-layer activations into the memory slots and simultaneously computes a content summary $\bar{\mathbf{h}}_t$, which is combined with the current timestep $t$ to form the updated conditioning feature $\mathbf{t}_{\mathrm{cond}}$. The Updater then integrates this new context with the existing state via gated recurrence, completing the recurrent loop.

\subsubsection{Shared Time Conditioner and AdaRMSNorm}
\label{sec:time-and-norm}

All three MetaState modules require a shared conditioning signal that captures both the diffusion timestep and the current content of the sequence. A pure timestep embedding is insufficient because the optimal modulation depends on which tokens have already been revealed at each step. We therefore construct a shared time conditioner from a sinusoidal embedding~\citep{vaswani2017attention} followed by an MLP, with a zero-gated content residual:
\begin{align*}
\mathbf{t}_{\mathrm{cond}} = \mathrm{MLP}\!\left(\mathrm{sinusoidal}(t)\right) \;+\; \boldsymbol{\alpha}_g \odot W_c\!\left(\mathrm{RMSNorm}(\bar{\mathbf{h}}_t)\right) \in \mathbb{R}^{d_c},
\end{align*}
where $d_c$ is the output dimension of the time conditioner, $\bar{\mathbf{h}}_t \in \mathbb{R}^{d_m}$ is the mean-pooled content summary derived from down-projected backbone hidden states (computed by the Mixer before cross-attention, \S\ref{sec:mixer}), $W_c$ projects to time dimension $d_c$, and $\boldsymbol{\alpha}_g \in \mathbb{R}^{d_c}$ is a learnable per-channel gate. The gate is zero-initialized to ensure that the conditioner begins as a pure timestep function and only gradually incorporates content-aware modulation as training proceeds, avoiding unstable early-stage interactions.

The conditioning feature $\mathbf{t}_{\mathrm{cond}}$ is consumed by the normalization layer through AdaRMSNorm, an adaptive variant of RMSNorm~\citep{zhang2019root}. Each AdaRMSNorm layer includes a modulation projection $W_{\mathrm{mod}}$ that produces per-channel scale and shift parameters from the conditioning:
\begin{align*}
[\boldsymbol{\gamma},\boldsymbol{\beta}] = W_{\mathrm{mod}}(\mathbf{t}_{\mathrm{cond}}), \quad
\mathcal{N}(\mathbf{x},t) = (1 + \boldsymbol{\gamma})\odot \mathrm{RMSNorm}(\mathbf{x}) + \boldsymbol{\beta}.
\end{align*}
At initialization, AdaRMSNorm reduces to standard RMSNorm, preserving the pretrained behavior of any layer it wraps. We also define a zero-bridge variant $\mathcal{N}_0(\mathbf{x},t) = \boldsymbol{\gamma}\odot \mathrm{RMSNorm}(\mathbf{x}) + \boldsymbol{\beta}$, which outputs nearly $\mathbf{0}$ at initialization. This serves as a zero bridge in the Injector~(\S\ref{sec:injector}), ensuring that the augmented model begins as the unmodified backbone.

\subsubsection{MetaState Mixer}
\label{sec:mixer}

The Mixer is designed to convert the variable-length backbone activations $\mathbf{h}_t\in\mathbb{R}^{N\times D}$ into a fixed-size representation of $M$ slots, where each slot should capture distinct aspects of the computation rather than collapse into a single pooled summary. We therefore use cross-attention with the state slots as queries, letting each slot selectively read the most informative tokens. To keep the module lightweight, the cross-attention operates in a $d_m$-dimensional bottleneck, and an up-projection recovers the full state dimension $D_s$. Before entering the bottleneck, a slot self-attention layer $\mathrm{Attn}_{\mathrm{GQA}}$ with plain RMSNorm enables inter-slot coordination in the full $D_s$ space, encouraging different slots to specialize and avoid redundant reads. Both the state and the hidden representation are then down-projected and time-conditioned within the bottleneck:
\begin{align*}
\mathbf{s}^b_t = \mathcal{N}(W^s_{\downarrow}\,\mathbf{s}_t, \;t) \in \mathbb{R}^{M \times d_m}, \quad
\mathbf{h}^b_t = \mathcal{N}(W^h_{\downarrow}\,\mathbf{h}_t, \;t) \in \mathbb{R}^{N \times d_m}.
\end{align*}
Before cross-attention, the Mixer computes a content summary $\bar{\mathbf{h}}_t = \mathrm{MeanPool}(W^h_\downarrow \mathbf{h}_t)$ and passes it to the time conditioner~(\S\ref{sec:time-and-norm}), so that subsequent normalization layers can adapt to the current sequence content.
Cross-attention is then computed with $\mathbf{s}^b_t$ as queries and $\mathbf{h}^b_t$ as keys/values, yielding $\mathbf{a}^b_t = \mathrm{CrossAttn}_{\mathrm{GQA}}(\mathbf{s}^b_t, \mathbf{h}^b_t) \in \mathbb{R}^{M \times d_m}$.
An FFN with AdaRMSNorm is then applied, followed by an up-projection to yield the Mixer output $\mathbf{c}_t \in \mathbb{R}^{M\times D_s}$.

\subsubsection{MetaState Updater}
\label{sec:updater}

The Updater must retain information accumulated over earlier denoising steps while incorporating new context from the current step. A time-conditioned GRU~\citep{dey2017gate} addresses this trade-off directly: its learned update gate provides a per-dimension interpolation between the existing state and a candidate update. Given the current state $\mathbf{s}_t\in\mathbb{R}^{M\times D_s}$ and the Mixer output $\mathbf{c}_t\in\mathbb{R}^{M\times D_s}$, both inputs are first normalized with time conditioning. A learnable slot identity embedding $\mathbf{e}_\mathrm{slot} \in \mathbb{R}^{M \times D_s}$ is added to the state before normalization so that each slot can learn distinct retention and update behaviors. The complete update rule is:
\begin{align*}
&\bar{\mathbf{s}}_t = \mathcal{N}(\mathbf{s}_t + \mathbf{e}_\mathrm{slot},\;t), \quad
\bar{\mathbf{c}}_t = \mathcal{N}(\mathbf{c}_t,\;t), \\
&\mathbf{z}_t,\mathbf{r}_t = \sigma\!\left(W_g\!\left([\bar{\mathbf{s}}_t \,\|\, \bar{\mathbf{c}}_t]\right)\right), \quad
\tilde{\mathbf{s}}_t = \tanh\!\left(W_{\tilde{s}}\!\left([\mathbf{r}_t\odot \bar{\mathbf{s}}_t \,\|\, \bar{\mathbf{c}}_t]\right)\right), \\
&\mathbf{s}_{t-1} = (1-\mathbf{z}_t)\odot \mathbf{s}_t + \mathbf{z}_t\odot \tilde{\mathbf{s}}_t.
\end{align*}
The update gate $\mathbf{z}_t$ controls how much of each dimension is overwritten, while the reset gate $\mathbf{r}_t$ determines how much of the previous state influences the candidate $\tilde{\mathbf{s}}_t$. The final interpolation ensures a smooth transition between retaining old information and integrating new context. Time modulation enters the GRU exclusively through the AdaRMSNorm layers on $\bar{\mathbf{s}}_t$ and $\bar{\mathbf{c}}_t$, which effectively provides timestep-dependent information.

\subsubsection{MetaState Injector}
\label{sec:injector}

The Injector should write the persistent state back into the backbone without disrupting its pretrained capabilities. We therefore realize it as an additive modulation of the input embeddings. Given embeddings $\mathbf{e}_t\in\mathbb{R}^{N\times D}$, we first down-project to a $d_b$-dimensional bottleneck: $\mathbf{x}^b_t = W^e_{\downarrow}\,\mathbf{e}_t \in \mathbb{R}^{N \times d_b}$. A self-attention layer $\mathrm{Attn}_{\mathrm{GQA}}$ with sinusoidal positional encoding and plain RMSNorm then enriches these representations with explicit positional context, enabling the subsequent cross-attention to route slot information to the appropriate sequence positions. The state is down-projected and normalized via $\mathbf{s}^b_t = \mathcal{N}(W^s_{\downarrow}\,\mathbf{s}_t, t)$, and cross-attention is computed with $\mathbf{x}^b_t$ as queries and $\mathbf{s}^b_t$ as keys/values. The output is added as a residual to $\mathbf{x}^b_t$. A subsequent FFN refines the fused representation, and a zero-bridge layer ($\mathcal{N}_0$) up-projects the result back to the full embedding dimension:
\begin{align*}
\mathbf{x}^b_t \leftarrow \mathbf{x}^b_t + \mathrm{FFN}\!\left(\mathcal{N}(\mathbf{x}^b_t, t)\right), \quad
\boldsymbol{\delta}_t = W_{\uparrow}\,\mathcal{N}_0(\mathbf{x}^b_t, t), \quad \tilde{\mathbf{e}}_t = \mathbf{e}_t + \boldsymbol{\delta}_t.
\end{align*}
Because $\mathcal{N}_0$ outputs near-zero at initialization, the modulation $\boldsymbol{\delta}_t$ vanishes at the start of training, ensuring that the augmented model begins as the unmodified backbone while still allowing all Injector parameters to receive gradients. The final modified embeddings $\tilde{\mathbf{e}}_t$ are fed back into the frozen backbone for the current denoising step.

\subsection{Training: $K$-Step Iterative Unrolling}
\label{sec:training}

Standard masked diffusion training samples a single random timestep $t$ per example and optimizes a single-step denoising objective. This approach is inadequate for MetaState, whose persistent state forms a recurrent chain across the full denoising trajectory (\S\ref{sec:metastate-overview}): the modules must learn what information to write into the state, what to retain across steps, and how to adapt the gating behavior across the denoising trajectory. We therefore adopt a multi-step unrolling pipeline with backpropagation through time (BPTT)~\citep{werbos2002backpropagation, gers2002learning} along the state trajectory.

\textbf{Training Pipeline.} Starting from a fully masked input, a warmup forward pass initializes the recurrent state without computing loss. A complete reveal trajectory is then pre-sampled to partition all $N_m$ maskable positions into $K$ batches. At each unrolling step, the model first predicts on the current masked input, and a batch of positions is then revealed via teacher forcing.
Let $\mathcal{M}\subseteq\{1,\dots,N\}$ denote the set of maskable positions, and let $N_m = |\mathcal{M}|$ denote the total number of such positions. At unrolling step $k$, let $\mathcal{M}_k$ denote the set of positions that remain masked, with $\mathcal{M}_1 = \mathcal{M}$ and $\mathcal{M}_k \subset \mathcal{M}_{k-1}$. Let $\mathcal{R}_k \subset \mathcal{M}_k$ denote the set of $n_k$ positions revealed at step $k$. The sequence $\mathbf{x}^{(k+1)}$ is then obtained by revealing the ground-truth tokens at the positions in $\mathcal{R}_k$, and the masked set is updated as $\mathcal{M}_{k+1} = \mathcal{M}_k \setminus \mathcal{R}_k$. The complete training pipeline is summarized in Algorithm~\ref{alg:metastate-training} in the Appendix.

\textbf{Dirichlet trajectory.}
We pre-sample reveal counts $\mathbf{n} = (n_1,\ldots,n_K)$ from a symmetric Dirichlet--Multinomial distribution over the $K$ denoising steps, which partitions the $N_m$ maskable positions into step-wise reveal budgets. A random permutation of maskable positions determines the reveal order, and step $k$ reveals the next $n_k$ positions in that order. The timestep at step $k$ is defined as a normalized masked ratio $t^{(k)} = |\mathcal{M}_k| / N_m \in [0, 1]$.

\subsubsection{Loss Function}
\label{sec:training-loss}

At each unrolling step \(k\), the objective interpolates between dense supervision over all positions that remain masked and focused supervision over the subset scheduled for revelation:
\[
\mathcal{L}_k
=
w_k \Big[
\lambda_d \,\ell_k^{\mathrm{dense}}
+
(1-\lambda_d)\,\ell_k^{\mathrm{reveal}}
\Big],
\qquad
w_k = \frac{n_k}{N_m},
\]
where \(n_k = |\mathcal{R}_k|\) is the number of positions revealed at step \(k\), and \(N_m\) is the total number of maskable positions. The dense and reveal losses are defined as
\[
\ell_k^{\mathrm{dense}}
=
\frac{1}{|\mathcal{M}_k|}
\sum_{i \in \mathcal{M}_k}
-\log p_\theta\!\left(x_0^{(i)} \mid \mathbf{x}^{(k)}\right),
\qquad
\ell_k^{\mathrm{reveal}}
=
\frac{1}{n_k}
\sum_{i \in \mathcal{R}_k}
-\log p_\theta\!\left(x_0^{(i)} \mid \mathbf{x}^{(k)}\right),
\]
where \(\mathcal{M}_k\) denotes the set of positions still masked at step \(k\), and \(\mathcal{R}_k \subseteq \mathcal{M}_k\) denotes the subset selected for revelation at that step.

We also include a regularization term to penalize per-slot norms that exceed a threshold $\tau$:
\begin{align*}
\mathcal{R}eg_s = \frac{\lambda_s}{KM}\sum_{k=1}^K \sum_{m=1}^M [\max(0, \|s_m^{(k)}\|_2 - \tau)]^2,
\end{align*}
where $\tau$ leaves the state free below the threshold, and $\lambda_s$ is the regularization weight. The total loss is the sum of per-step losses $\sum_{k=1}^{K}\mathcal{L}_k$ and the regularization term $\mathcal{R}eg_s$.

\section{Experiments}

\definecolor{deltaGreen}{HTML}{4F9F4F}
\newcommand{\deltac}[1]{\textcolor{deltaGreen}{#1}}

\begin{table*}[t]
\caption{Main results on 4 different benchmarks (generation length 256, block size 32, dual cache). MATH\,{=}\,MATH-500, HE\,{=}\,HumanEval. $\Delta$ denotes improvement over the corresponding baseline. \textbf{Bold} marks the best result per column within each backbone group.}
\label{tab:main-results}
\centering
\small
\begin{tabular}{@{}l cccc @{\hspace{14pt}} cccc @{\hspace{14pt}} c@{}}
\toprule
\toprule
& \multicolumn{4}{c}{\textit{Dream Backbone (7B)}} & \multicolumn{4}{c}{\textit{LLaDA Backbone (8B)}} & \\
\cmidrule(lr){2-5} \cmidrule(lr){6-9}
Model & GSM8K & MATH & HE & MBPP & GSM8K & MATH & HE & MBPP & \textbf{Avg.} \\
\midrule
Base & 73.7 & 37.6 & 54.9 & 52.6 & 67.4 & 28.8 & 33.5 & 25.6 & \cellcolor{orange!8}46.8 \\
\quad + \textbf{MetaState} & \textbf{76.7} & \textbf{46.4} & \textbf{59.2} & \textbf{53.6} & \textbf{77.9} & \textbf{37.0} & \textbf{39.6} & \textbf{33.0} & \cellcolor{orange!8}\textbf{52.9} \\
\quad {\footnotesize $\Delta$ vs.\ Base} & \deltac{+3.0} & \deltac{+8.8} & \deltac{+4.3} & \deltac{+1.0} & \deltac{+10.5} & \deltac{+8.2} & \deltac{+6.1} & \deltac{+7.4} & \cellcolor{orange!8}\deltac{+6.2} \\
\midrule
Instruct & 74.8 & 45.0 & 56.1 & 51.0 & 78.5 & 36.8 & 37.2 & 26.0 & \cellcolor{orange!8}50.7 \\
\quad + \textbf{MetaState} & \textbf{78.1} & \textbf{46.6} & \textbf{59.8} & \textbf{55.0} & \textbf{79.5} & \textbf{37.8} & \textbf{39.6} & \textbf{32.2} & \cellcolor{orange!8}\textbf{53.6} \\
\quad {\footnotesize $\Delta$ vs.\ Instruct} & \deltac{+3.3} & \deltac{+1.6} & \deltac{+3.7} & \deltac{+4.0} & \deltac{+1.0} & \deltac{+1.0} & \deltac{+2.4} & \deltac{+6.2} & \cellcolor{orange!8}\deltac{+2.9} \\
\bottomrule
\bottomrule
\vspace{-2.5em}
\end{tabular}
\end{table*}

\subsection{Experimental Settings}

\textbf{Models and Datasets.}
We apply MetaState to two discrete diffusion LLM families, each in Base and Instruct variants: LLaDA-Instruct-8B / LLaDA-Base-8B~\citep{nie2025large} and Dream-v0-Instruct-7B / Dream-v0-Base-7B~\citep{ye2025dream}. To isolate the effect of the recurrent design, all backbone parameters are frozen throughout training. Only the MetaState components (Mixer, Updater, Injector, and the time-conditioning module) are trained, amounting to approximately 0.6\% of each backbone. We train on 50{,}000 sequences sampled from the T\"{u}lu-3 SFT mixture (allenai/tulu-3-sft-mixture)~\citep{lambert2024tulu3}, using each model's native tokenizer and chat template with a maximum sequence length of 1024.

\textbf{Evaluation Benchmarks.}
We evaluate on four reasoning benchmarks: GSM8K~\citep{cobbe2021gsm8k} (5-shot) and MATH-500~\citep{lewkowycz2022solving} (4-shot) for mathematical reasoning, and HumanEval~\citep{chen2021codex} (0-shot) and MBPP~\citep{austin2021program} (3-shot) for code generation. For code benchmarks, accuracy refers to Pass@1 measured by functional correctness on unit tests. Following standard dLLM practice, the generation length is 256 with a block size of 32. Additional decoding-configuration results are shown in Appendix~\S\ref{sec:decoding-robustness}. All evaluations use the KV-cache and parallel decoding of Fast-dLLM~\citep{wu2025fast}.

\subsection{Main Results}

Table~\ref{tab:main-results} compares MetaState against both Base and Instruct backbones on four reasoning benchmarks. MetaState consistently improves accuracy over both Base and Instruct variants across all benchmarks and both dLLM families.
On Dream, MetaState outperforms Dream-Base on every benchmark, with gains of $+3.0$ on GSM8K, $+8.8$ on MATH-500, $+4.3$ on HumanEval, and $+1.0$ on MBPP. The same trend holds against the stronger Dream-Instruct baseline, where MetaState improves GSM8K by $+3.3$, MATH-500 by $+1.6$, HumanEval by $+3.7$, and MBPP by $+4.0$.
LLaDA exhibits the same pattern at a larger scale. Relative to LLaDA-Base, MetaState yields gains of $+10.5$ on GSM8K, $+8.2$ on MATH-500, $+6.1$ on HumanEval, and $+7.4$ on MBPP. Relative to LLaDA-Instruct, the improvements are smaller but remain consistent, with gains of $+1.0$ on GSM8K, $+1.0$ on MATH-500, $+2.4$ on HumanEval, and $+6.2$ on MBPP. The smaller margins over Instruct backbones are consistent with lower recoverable cross-step loss after instruction tuning; MetaState does not update backbone weights and instead changes the augmented system through an input-side recurrent memory interface.

These results suggest that persistent working memory is particularly beneficial for tasks that require information to remain stable across long denoising trajectories. In mathematical reasoning, the model must retain intermediate computations and partial conclusions until the final answer is formed. In code generation, it must maintain global structural constraints such as variable scope, control flow, and program-level consistency over many lines of code. Both settings are vulnerable to cross-step drift, and the consistent gains of MetaState across all four benchmarks and two architecturally distinct dLLM families support the view that its benefits arise from mitigating the Information Island issue.
We further evaluate MetaState on the newer RL-optimized LLaDA 1.5 backbone in Appendix~\ref{sec:llada15-results}; MetaState improves all four evaluated metrics under the same protocol, supporting transfer beyond the original checkpoints.

\subsection{Compatibility with Soft Diffusion}

\begin{table*}[t]
\caption{Compatibility with Soft Diffusion~\citep{zhu2025latent}.
MetaState and Soft Diffusion target different levels of the pipeline and can be combined.
$\dagger$\,denotes\,`+\,Soft Diffusion'. \textbf{Bold} marks the best result per column.
Hyperparameter details are provided in Appendix~\ref{sec:soft-diffusion-details}.}
\label{tab:soft-diffusion}
\centering
\small
\begin{tabular}{@{}l cccc @{\hspace{10pt}} cccc @{\hspace{10pt}} c@{}}
\toprule
\toprule
& \multicolumn{4}{c}{\textit{Dream-Instruct (7B)}}
& \multicolumn{4}{c}{\textit{LLaDA-Instruct (8B)}} & \\
\cmidrule(lr){2-5} \cmidrule(lr){6-9}
Method & GSM8K & MATH & HE & MBPP
       & GSM8K & MATH & HE & MBPP & \textbf{Avg.} \\
\midrule
Instruct baseline
  & 74.8 & 45.0 & 56.1 & 51.0
  & 78.5 & 36.8 & 37.2 & 26.0 & \cellcolor{orange!8}50.7 \\
\quad + Soft Diffusion
  & 79.1 & 45.8 & \textbf{60.4} & 52.0
  & 78.7 & \textbf{37.8} & 37.8 & 29.0 & \cellcolor{orange!8}52.6 \\
\midrule
\quad + MetaState
  & 78.1 & \textbf{46.6} & 59.8 & 55.0
  & 79.5 & \textbf{37.8} & 39.6 & 32.2 & \cellcolor{orange!8}53.6 \\
\quad + MetaState$^\dagger$
  & \textbf{79.4} & 46.4 & 59.2 & \textbf{55.6}
  & \textbf{80.3} & 37.6 & \textbf{40.9} & \textbf{32.4} & \cellcolor{orange!8}\textbf{54.0} \\
\bottomrule
\bottomrule
\vspace{-2.5em}
\end{tabular}
\end{table*}

Both MetaState and Soft Diffusion~\citep{zhu2025latent} improve dLLM decoding but through orthogonal mechanisms. Soft Diffusion directly modifies the original discrete decoding path by replacing hard masked token positions with probability-weighted embedding mixtures. MetaState, in contrast, leaves the discrete path unchanged and instead introduces a parallel persistent memory path that carries continuous information. In this sense, Soft Diffusion refines how token representations are formed within each step, whereas MetaState augments the decoding process with an additional cross-step information channel. Because one modifies the discrete path itself and the other adds a separate recurrent pathway alongside it, the two methods are naturally orthogonal and can be combined. Table~\ref{tab:soft-diffusion} evaluates this combination on the Instruct variants of both backbones.

Comparing MetaState and Soft Diffusion individually (rows~2 and 3 of Table~\ref{tab:soft-diffusion}), MetaState outperforms Soft Diffusion on most benchmarks. Applying Soft Diffusion on top of MetaState brings further improvements, and the combination achieves the strongest overall results, including the best GSM8K accuracy of $80.3$ and $79.4$ on LLaDA and Dream. These results support the view that the two methods are complementary. The only exception is Dream-HumanEval, where MetaState + Soft Diffusion ($59.2$) underperforms both MetaState alone ($59.8$) and Soft Diffusion alone ($60.4$). We attribute this to two factors. First, HumanEval contains only 164 problems, so small differences in accuracy are inherently noisy. Second, the Injector is trained with pure mask embeddings as input, whereas Soft Diffusion replaces them with probability-weighted embedding mixtures, introducing an input distribution shift that may interfere with the Injector's additive modulation. Further details on Soft Diffusion hyperparameters are provided in Appendix~\ref{sec:soft-diffusion-details}.

\subsection{Ablation Studies}

Table~\ref{tab:ablation} presents ablation results for each MetaState component on both Dream-Instruct and LLaDA-Instruct under the same 50k-sample training setup.

\definecolor{tblgray}{rgb}{0.5, 0.5, 0.5}
\begin{table}[t]
\centering
\caption{Ablation studies. Each row modifies one component while keeping the rest of MetaState intact. MLP variants are parameter-matched. MATH\,{=}\,MATH-500, HE\,{=}\,HumanEval. \textbf{Bold}: full model. \textcolor{tblgray}{Gray}: backbone without MetaState (from Table~\ref{tab:main-results}).}
\label{tab:ablation}
\small
\begin{tabular}{@{}l cccc @{\hspace{14pt}} cccc @{\hspace{14pt}} c@{}}
\toprule
\toprule
& \multicolumn{4}{c}{\textit{Dream-Instruct (7B)}} & \multicolumn{4}{c}{\textit{LLaDA-Instruct (8B)}} & \\
\cmidrule(lr){2-5} \cmidrule(lr){6-9}
Model Variants & GSM8K & MATH & HE & MBPP & GSM8K & MATH & HE & MBPP & \textbf{Avg.} \\
\midrule
\textcolor{tblgray}{\textit{Backbone only}} & \textcolor{tblgray}{74.8} & \textcolor{tblgray}{45.0} & \textcolor{tblgray}{56.1} & \textcolor{tblgray}{51.0} & \textcolor{tblgray}{78.5} & \textcolor{tblgray}{36.8} & \textcolor{tblgray}{37.2} & \textcolor{tblgray}{26.0} & \cellcolor{orange!8}\textcolor{tblgray}{50.7} \\
\textbf{MetaState (full)} & \textbf{78.1} & \textbf{46.6} & \textbf{59.8} & \textbf{55.0} & \textbf{79.5} & \textbf{37.8} & \textbf{39.6} & \textbf{32.2} & \cellcolor{orange!8}\textbf{53.6} \\
\midrule
w/o recurrence      & 77.6 & 45.0 & 61.0 & 53.4 & 62.6 & 32.6 & 23.8 & 24.6 & \cellcolor{orange!8}47.6 \\
w/o BPTT            & 76.7 & 43.4 & 59.2 & 53.2 & 78.2 & 34.4 & 38.4 & 33.0 & \cellcolor{orange!8}52.1 \\
w/o time cond.      & 77.2 & 44.8 & 57.9 & 54.6 & 78.9 & 36.2 & 39.0 & 31.8 & \cellcolor{orange!8}52.6 \\
MLP Injector        & 77.2 & 41.2 & 54.3 & 52.2 & 77.7 & 36.2 & 38.4 & 30.8 & \cellcolor{orange!8}51.0 \\
MLP Mixer           & 76.9 & 42.4 & 56.7 & 54.2 & 74.4 & 33.8 & 33.5 & 32.2 & \cellcolor{orange!8}50.5 \\

\bottomrule
\bottomrule
\vspace{-2.5em}
\end{tabular}
\end{table}

\textbf{Recurrence.}\; Zeroing the previous state at every denoising step (w/o recurrence) has asymmetric effects across backbones: on LLaDA, performance drops substantially below the backbone-only baseline ($-15.9$ on GSM8K and $-13.4$ on HumanEval), indicating that the Injector without recurrent state becomes harmful. On Dream, in contrast, the effect is minimal, and the variant still remains above the Dream backbone on three of four benchmarks. This asymmetry suggests that Dream retains stronger per-step coherence during denoising, whereas LLaDA depends more heavily on the recurrent memory channel to preserve information across steps. Detaching the state between denoising steps (w/o BPTT), which preserves accumulation but blocks gradient flow through the unrolled trajectory, consistently degrades both backbones, with the largest drops on MATH-500 ($-3.2$ for Dream, $-3.4$ for LLaDA), showing that BPTT is important for learning useful state dynamics.

\textbf{Architecture.}\; Replacing the attention-based Injector with a parameter-matched MLP (MLP Injector) that flattens the slot state and broadcasts a uniform bias removes position-aware injection while preserving a path from memory to the token representations. This leads to clear degradation on both backbones, with particularly large drops on Dream ($-5.5$ on HumanEval and $-5.4$ on MATH-500). Replacing the attention-based Mixer with a parameter-matched MLP (MLP Mixer) that average-pools backbone hidden states before projection likewise hurts performance consistently, and is especially damaging on LLaDA ($-5.1$ on GSM8K and $-6.1$ on HumanEval). Together, these results show that both token-selective reading in the Mixer and position-aware writing in the Injector are important. Removing explicit time conditioning by zeroing the SharedTimeConditioner (w/o time cond.) produces the smallest drops across benchmarks, suggesting that while timestep information is beneficial, the recurrent gating and attention mechanisms can partially infer denoising progress even without explicit conditioning.

\section{Conclusion}

We presented MetaState, a lightweight recurrent augmentation together with a dedicated training pipeline that equips frozen discrete diffusion LLM backbones with a persistent, fixed-size working memory across denoising steps, thereby addressing the Information Island issue. On frozen LLaDA-8B and Dream-7B backbones, MetaState only adds approximately $0.6\%$ trainable parameters and yields consistent improvements on both mathematical reasoning and code generation benchmarks, demonstrating that persistent cross-step working memory is an effective mechanism for improving reasoning performance in dLLMs.
\section{Acknowledgments}
This material is based upon work supported by the National Science Foundation under grant No.2229876 and is supported in part by funds provided by the National Science Foundation, by the Department of Homeland Security, and by IBM. Any opinions, findings, and conclusions or recommendations expressed in this material are those of the author(s) and do not necessarily reflect the views of the National Science Foundation or its federal agency and industry partners.

\bibliographystyle{colm2026_conference}
\bibliography{colm2026_conference}

\newpage
\appendix
\section{Appendix}

\subsection{The Information Island Issue}
\label{sec:info-island}

The Information Island issue arises from the representational gap between the continuous hidden state $\mathbf{h}_t$ computed within each denoising step and the discrete sequence $\mathbf{x}_{t-1}$ passed to the next step. This gap is inherent to the Markovian formulation of dLLM decoding: each inter-step transition conditions solely on the current discrete state $\mathbf{x}_t$, while the continuous representation $\mathbf{h}_t$ is not carried forward explicitly. On one side, $\mathbf{h}_t$ encodes substantially richer information than what the discrete tokens can carry: beyond token-level predictive distributions, it captures long-range dependencies, partial reasoning information, and global structural constraints over the sequence. On the other side, the next-step input $\mathbf{x}_{t-1} = \mathcal{S}(\mathbf{h}_t)$ is produced by the sampling-and-remasking operator $\mathcal{S}$, which retains only discrete token identities at a sparse subset of positions and discards all remaining continuous context embedded in $\mathbf{h}_t$. This persistent gap at every transition along the denoising trajectory is what we term the Information Island issue.

This bottleneck degrades dLLMs in several related ways. First, useful information computed at one step cannot be directly reused at the next step, and the subsequent step can only re-derive it from a sparse, partially masked sequence that carries no trace of the earlier computation. Second, because this re-derivation occurs under changing noise levels and mask patterns, the trajectory can drift: information that is correct at one step may be weakened, overwritten, or inconsistently re-derived at later steps. Third, the problem is especially severe for reasoning. Tasks such as multi-step mathematics and code generation require the model to make use of intermediate computations. Without an explicit cross-step memory mechanism, these intermediate results are repeatedly exposed to the lossy discrete interface, making it difficult for them to make a useful contribution to later steps.

\begin{figure}[ht]
    \centering
    \vspace{-0.4em}
    \includegraphics[width=\linewidth]{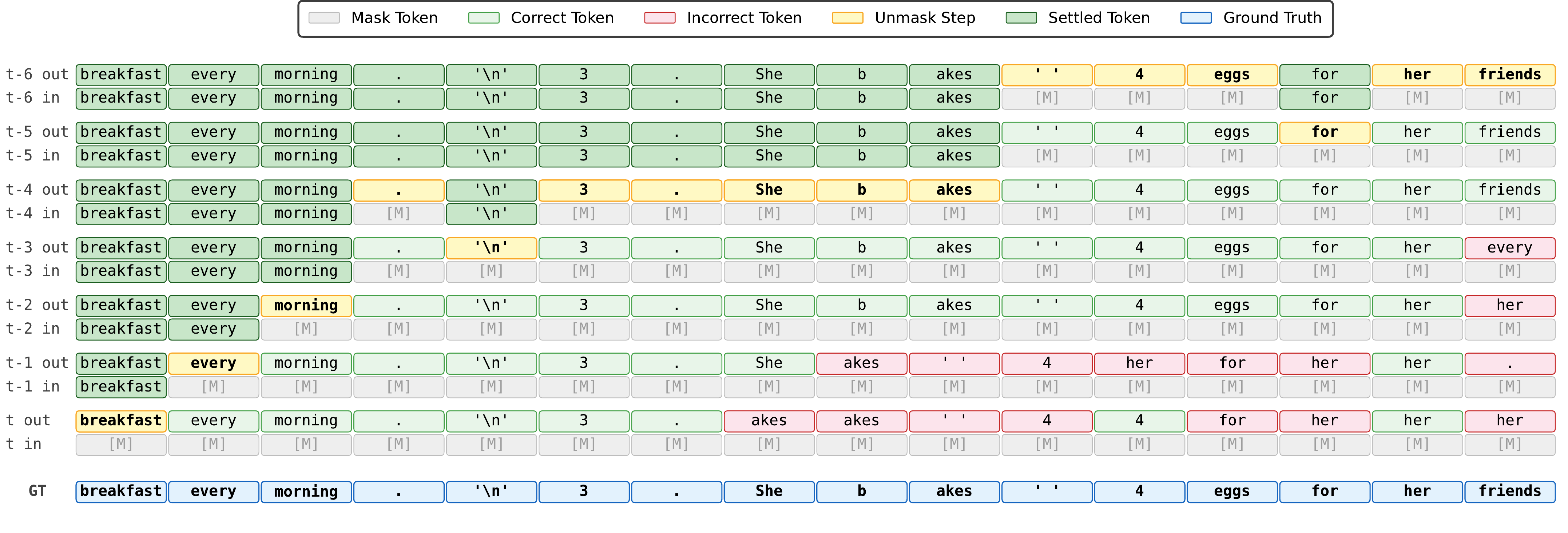}
    \vspace{-2.4em}
    \caption{A step-by-step denoising trajectory. Each denoising step shows both the full argmax model prediction (\texttt{out}) and the remasked sequence (\texttt{in}) that is actually passed to the next step.}
    \label{fig:appendix_island_case}
    \vspace{-0.8em}
\end{figure}

\textbf{A step-by-step denoising example.}
Figure~\ref{fig:appendix_island_case} makes this failure mode concrete by showing, at each denoising step, both (i) the full decoded prediction before remasking denoted by $\hat{\mathbf{x}}_{t}$, and (ii) the remasked sequence input $\mathbf{x}_{t-1}$ that is actually passed to the next step. The key observation is that correct information can already appear in $\hat{\mathbf{x}}_t$ several steps before generation is finalized, yet much of it is lost after remasking and never reaches subsequent steps. Each subsequent step receives only a partial discrete snapshot and must re-derive the missing relations from scratch. In the example, some correct tokens or reasoning fragments appear early, but because they are not preserved across the discrete interface, later steps fail to reuse them and may instead drift toward an inconsistent continuation.

\begin{wrapfigure}{t}{0.6\textwidth}
    \centering
    \vspace{-\intextsep}
\includegraphics[width=\linewidth]{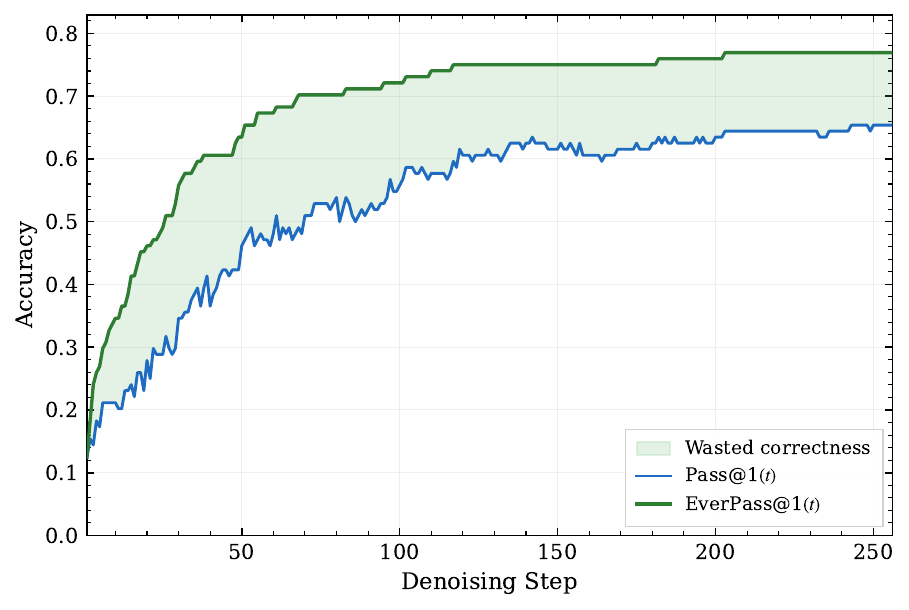}
    \vspace{-2.4em}
    \caption{Comparison of \textsc{Pass@1}$(t)$ and \textsc{EverPass@1}$(t)$ across denoising steps on GSM8K with LLaDA-Instruct-8B.}
    \label{fig:appendix_everpass}
    \vspace{-\intextsep}
\end{wrapfigure}

\textbf{Trajectory-level evidence from \textsc{Pass@1} and \textsc{EverPass@1}.}
The Information Island issue not only discards useful intermediate representations, but can also degrade final generation quality, as correct tokens produced at earlier steps may be overwritten with incorrect ones after passing through the discrete sampling-and-remasking interface. Following recent analyses of temporal dynamics in dLLMs~\citep{wang2025time}, Figure~\ref{fig:appendix_everpass} compares the \textsc{Pass@1} with \textsc{EverPass@1}, where \textsc{EverPass@1} counts an example as successful if any intermediate full prediction along the denoising trajectory is correct. 
\vspace{1em}
On our test, \textsc{EverPass@1} remains significantly higher than the \textsc{Pass@1}, indicating that many correct results are already present at intermediate denoising steps but are lost through the subsequent discrete remasking operation. In other words, the model often ``knows'' the right answer somewhere along the trajectory, yet fails to preserve that information through subsequent sampling and remasking. Together with the case study in Figure~\ref{fig:appendix_island_case}, these statistics show that the Information Island issue is a common failure mode in discrete diffusion language models.

These observations motivate MetaState, which introduces a persistent continuous working memory that carries useful intermediate information across steps and tackles the problems mentioned above.

\subsection{Additional Trajectory Evidence}
\label{sec:additional-trajectory-evidence}

We further compare the trajectory-level \textsc{Pass@1} and \textsc{EverPass@1} diagnostics before and after adding MetaState on a subset of GSM8K. This diagnostic tests whether MetaState helps correct intermediate predictions survive the sampling-and-remasking interface. Table~\ref{tab:rebuttal-pass-everpass} shows that MetaState improves \textsc{Pass@1} throughout the trajectory and reduces the wasted-correctness gap. At the final step, both systems reach the same \textsc{EverPass@1} of 79.00, while final \textsc{Pass@1} improves from 63.00 to 66.00. This pattern indicates that the augmented model does not merely make more answers reachable somewhere along the trajectory; it preserves more of the correct information that is already reachable so that it survives into the final output.

\begin{table*}[t]
\centering
\small
\caption{GSM8K trajectory diagnostic for LLaDA-Instruct. \textsc{EverPass@1} counts an example as correct if any intermediate full prediction along the denoising trajectory is correct. The gap is \textsc{EverPass@1} minus \textsc{Pass@1}.}
\label{tab:rebuttal-pass-everpass}
\begin{tabular}{@{}l c c c c c c c@{}}
\toprule
 Steps & Orig. PASS & Our PASS & Orig. EVER & Our EVER & gain & Orig. gap & Our gap \\
\midrule
 1--256 & 49.58 & 54.80 & 65.88 & 69.05 & +5.23 & 16.30 & 14.25 \\
 256 & 63.00 & 66.00 & 79.00 & 79.00 & +3.00 & 16.00 & 13.00 \\
 1--85 & 31.59 & 40.42 & 45.96 & 54.84 & +8.84 & 14.38 & 14.41 \\
 86--170 & 56.34 & 59.87 & 73.89 & 74.95 & +3.53 & 17.55 & 15.08 \\
 171--256 & 60.67 & 64.01 & 77.65 & 77.28 & +3.34 & 16.98 & 13.27 \\
\bottomrule
\end{tabular}
\end{table*}

\subsection{State and Memory Diagnostics}
\label{sec:state-memory-diagnostics}

We next inspect the recurrent state itself. Table~\ref{tab:rebuttal-state-memory} shows that consecutive-step hidden-state cosine in the original backbone is nearly identical for Dream and LLaDA, so this raw coherence statistic does not explain the backbone asymmetry. In contrast, MetaState's state norm, slot norm, and state variance are consistently larger on LLaDA. This indicates that the recurrent interface performs more active cross-step work on LLaDA, which is consistent with the stronger effect of recurrence-related ablations on that backbone.

\begin{table}[t]
\centering
\small
\caption{GSM8K state and memory diagnostics. Two 128-example subsets show the same pattern: the original consecutive-step hidden-state cosine is nearly identical across backbones, while MetaState memory activity is larger and more variable on LLaDA.}
\label{tab:rebuttal-state-memory}
\begin{tabular}{@{}l l c c@{}}
\toprule
Subset & Metric & Dream & LLaDA \\
\midrule
1 & Original response-step cosine, mean & 0.9749 & 0.9775 \\
2 & Original response-step cosine, mean & 0.9742 & 0.9769 \\
1 & MetaState state Frobenius norm, mean & 31.42 & 52.04 \\
2 & MetaState state Frobenius norm, mean & 31.21 & 51.98 \\
1 & MetaState mean slot norm, mean & 3.93 & 6.50 \\
2 & MetaState mean slot norm, mean & 3.90 & 6.50 \\
1 & State Frobenius pooled variance & 24.11 & 76.43 \\
2 & State Frobenius pooled variance & 24.27 & 77.29 \\
\bottomrule
\end{tabular}
\end{table}

We also run a numeric semantic retention probe on the incoming memory state received by the Injector before it writes into the backbone input embeddings. For each recorded state, a low-capacity linear scorer ranks candidate numbers against three targets: gold intermediate values, the model's final-answer candidate, and the gold final answer. The candidate pool includes question numbers, numbers already visible in the decoded response, target values, and magnitude-matched distractors. Table~\ref{tab:rebuttal-semantic-probe} reports mean reciprocal rank against the strongest among four controls: random ranking, constant-state features, question-number preference, and visible-text-number preference. MetaState exceeds the strongest control across both backbones and all target types, including the not-visible and not-in-question slices, showing that the state contains recoverable numeric reasoning information beyond the prompt and currently decoded text.

\begin{table*}[t]
\centering
\small
\caption{GSM8K numeric semantic retention probe. The probe reads the incoming MetaState memory before injection and ranks candidate numbers. Controls include random ranking, constant-state features, question-number preference, and visible-text-number preference.}
\label{tab:rebuttal-semantic-probe}
\begin{tabular}{@{}l l l c c c@{}}
\toprule
Backbone & Target & Slice & MetaState MRR & Best control MRR & Gap \\
\midrule
LLaDA & gold\_intermediate & all & 0.663 & 0.434 & +0.229 \\
LLaDA & gold\_intermediate & not\_visible & 0.588 & 0.363 & +0.225 \\
LLaDA & gold\_intermediate & not\_in\_question & 0.646 & 0.410 & +0.235 \\
LLaDA & model\_final & all & 0.308 & 0.220 & +0.088 \\
LLaDA & model\_final & not\_visible & 0.318 & 0.220 & +0.099 \\
LLaDA & model\_final & not\_in\_question & 0.311 & 0.220 & +0.092 \\
LLaDA & gold\_final & all & 0.327 & 0.220 & +0.107 \\
LLaDA & gold\_final & not\_visible & 0.326 & 0.219 & +0.106 \\
LLaDA & gold\_final & not\_in\_question & 0.327 & 0.219 & +0.107 \\
Dream & gold\_intermediate & all & 0.664 & 0.473 & +0.191 \\
Dream & gold\_intermediate & not\_visible & 0.546 & 0.349 & +0.197 \\
Dream & gold\_intermediate & not\_in\_question & 0.649 & 0.433 & +0.216 \\
Dream & model\_final & all & 0.303 & 0.217 & +0.086 \\
Dream & model\_final & not\_visible & 0.299 & 0.217 & +0.082 \\
Dream & model\_final & not\_in\_question & 0.305 & 0.217 & +0.087 \\
Dream & gold\_final & all & 0.315 & 0.217 & +0.098 \\
Dream & gold\_final & not\_visible & 0.315 & 0.216 & +0.098 \\
Dream & gold\_final & not\_in\_question & 0.317 & 0.216 & +0.101 \\
\bottomrule
\end{tabular}
\end{table*}

\newpage
\subsection{Pseudocode for MetaState}

Algorithm~\ref{alg:metastate-step} details the single denoising step procedure of MetaState, and Algorithm~\ref{alg:metastate-training} summarizes the full $K$-step iterative unrolling training procedure.

\vspace{-1em}
\begin{algorithm}[ht]
\caption{MetaState: Single Denoising Step}
\label{alg:metastate-step}
\begin{algorithmic}[1]
\Require Noisy sequence $\mathbf{x}_t$, persistent state $\mathbf{s}_t$ (or $\mathrm{None}$), previous time conditioning $\mathbf{t}_{\mathrm{cond}}$ (or $\mathrm{None}$), timestep $t$
\Ensure Logits $\hat{\mathbf{x}}_t$, updated state $\mathbf{s}_{t-1}$, time conditioning $\mathbf{t}_{\mathrm{cond}}$
\State $\mathbf{e}_t \gets \mathrm{Embed}(\mathbf{x}_t)$
\If{$\mathbf{s}_t = \mathrm{None}$} \Comment{Warmup step}
    \State $\mathbf{s}_t \gets \mathbf{s}_0$ \Comment{Learnable init}
    \State $\tilde{\mathbf{e}}_t \gets \mathbf{e}_t$ \Comment{Skip injection}
\Else
    \State $\tilde{\mathbf{e}}_t \gets \mathrm{Injector}(\mathbf{e}_t,\mathbf{s}_t,\mathbf{t}_{\mathrm{cond}})$ \Comment{Additive modulation}
\EndIf
\State $(\mathbf{h}_t,\hat{\mathbf{x}}_t) \gets p_\theta(\tilde{\mathbf{e}}_t)$ \Comment{Frozen backbone}
\State $\bar{\mathbf{h}}_t \gets \mathrm{MeanPool}(W^h_\downarrow\,\mathbf{h}_t)$ \Comment{Content summary}
\State $\mathbf{t}_{\mathrm{cond}} \gets \mathrm{TimeCond}(t,\bar{\mathbf{h}}_t)$ \Comment{Content-aware}
\State $\mathbf{c}_t \gets \mathrm{Mixer}(\mathbf{s}_t,\mathbf{h}_t,\mathbf{t}_{\mathrm{cond}})$ \Comment{Read from $\mathbf{h}_t$}
\State $\mathbf{s}_{t-1} \gets \mathrm{Updater}(\mathbf{s}_t,\mathbf{c}_t,\mathbf{t}_{\mathrm{cond}})$ \Comment{Update state}
\State \Return $(\hat{\mathbf{x}}_t,\mathbf{s}_{t-1},\mathbf{t}_{\mathrm{cond}})$ \Comment{For next step}
\end{algorithmic}
\end{algorithm}

\vspace{-1em}
\begin{algorithm}[ht]
\caption{MetaState Training}
\label{alg:metastate-training}
\begin{algorithmic}[1]
\Require Ground truth $\mathbf{x}_0$, maskable positions $\mathcal{M}$, $K$ steps
\State $\mathbf{x} \gets$ mask all positions in $\mathcal{M}$ with \texttt{[MASK]}
\State $(\mathrm{logits}, \mathbf{s}, \mathbf{t}_{\mathrm{cond}}) \gets \mathrm{Forward}(\mathbf{x}, \mathrm{state}{=}\mathrm{None}, \mathbf{t}_{\mathrm{cond}}{=}\mathrm{None}, t{=}1.0)$ \Comment{Warmup (Alg.~\ref{alg:metastate-step})}
\State Sample reveal counts $n_1,\ldots,n_K \sim \mathrm{Dir\text{-}Multi}$
\State Sample random reveal ranks for $\mathcal{M}$
\State $\mathcal{L} \gets 0$
\For{$k = 1$ to $K$}
    \State $t \gets |\text{still masked}| / N_m$ \Comment{Continuous timestep}
    \State $(\mathrm{logits}, \mathbf{s}, \mathbf{t}_{\mathrm{cond}}) \gets \mathrm{Forward}(\mathbf{x}, \mathbf{s}, \mathbf{t}_{\mathrm{cond}}, t)$ \Comment{Alg.~\ref{alg:metastate-step}}
    \State $\mathcal{L} \gets \mathcal{L} + \mathcal{L}_k(\mathrm{logits}, \mathbf{x}_0, \mathcal{M}_k, \mathcal{R}_k)$
    \State Reveal $n_k$ positions (teacher forcing), update $\mathbf{x}$
\EndFor
\State \Return $\mathcal{L} + \mathcal{R}eg_s$
\end{algorithmic}
\end{algorithm}

\FloatBarrier

\subsection{Experimental Details and Hyperparameters}

\subsubsection{Experimental Details.}
\label{sec:experimental-details}
We freeze the backbone and train only the MetaState recurrent components (Mixer, Updater, Injector, and SharedTimeConditioner) with AdamW ($\beta_1{=}0.9$, $\beta_2{=}0.95$), a peak learning rate of $2{\times}10^{-5}$ with cosine decay and 5\% linear warmup, and gradient clipping at max norm $1.0$.
All Metastate weights are initialized from a truncated normal distribution ($\sigma{=}0.02$), except from the zero-initialized ones.
We train for one epoch on 50{,}000 sequences from the T\"ulu-3 SFT mixture~\citep{lambert2024tulu3}, with a maximum sequence length of 1{,}024 tokens, using bfloat16 mixed precision with DeepSpeed ZeRO-1~\citep{rajbhandari2020zero} on two NVIDIA H200 GPUs.
Unless noted otherwise, all hyperparameters are shared across both backbones: $M{=}64$ memory slots, state dimension $D_s{=}1024$, bottleneck dimensions $d_m{=}d_b{=}768$, and unroll depth $K{=}4$.
Loss is computed only over the response portion, and prompt tokens are excluded from both masking and loss computation.
A hinge state-norm regularizer ($\lambda_s{=}1e{-}4$, threshold $\tau{=}1.0$) penalizes per-slot norms that exceed the threshold.

During evaluation, Dream performs a single bootstrap forward pass at $t{=}1.0$ with $\mathbf{s}{=}\mathrm{None}$ before denoising begins, initializing the recurrent state.
Chat-template application varies by backbone: LLaDA-Base evaluations omit the chat template entirely. LLaDA-Instruct applies it for GSM8K and MATH-500 but omits it for HumanEval and MBPP. All Dream variants (Base and Instruct) apply the backbone's native chat template.
Table~\ref{tab:training-hyperparams} lists the remaining inference hyperparameters.

\subsubsection{Architecture Hyperparameters.}
Table~\ref{tab:arch-hyperparams} summarizes the architectural configuration of all MetaState modules.

\begin{table}[ht]
\centering
\vspace{-1.6em}
\caption{Architecture hyperparameters for the recurrent modules. All symbols correspond to notation introduced in \S\ref{sec:method}.}
\label{tab:arch-hyperparams}
\begin{tabular}{@{}lll@{}}
\toprule
Symbol & Description & Value \\
\midrule
$M$ & Number of memory slots & $64$ \\
$D_s$ & State dimension per slot & $1024$ \\
$d_c$ & Time conditioner output dimension & $1024$ \\
$d_m$ & Mixer bottleneck dimension & $768$ \\
$d_b$ & Injector bottleneck dimension & $768$ \\
$n_q$ & Query heads (Mixer \& Injector) & $8$ \\
$n_{kv}$ & KV heads (GQA) & $4$ \\
--- & SwiGLU FFN expansion ratio & $2.0$ \\
$\boldsymbol{\alpha}_g$ & Content gate initialization & $\mathbf{0}$ (zero-init) \\
$\mathcal{N}_0$ & Zero-bridge soft bias & $10^{-3}$ \\
\bottomrule
\end{tabular}
\end{table}

\subsubsection{Training and Inference Hyperparameters.}
Table~\ref{tab:training-hyperparams} lists the optimization, unrolling, and inference settings shared across both LLaDA and Dream backbones.

\begin{table}[ht]
\centering
\caption{Training and inference hyperparameters (shared across LLaDA and Dream backbones).}
\label{tab:training-hyperparams}
\begin{tabular}{@{}ll@{}}
\toprule
Parameter & Value \\
\midrule
\multicolumn{2}{@{}l}{\textit{Optimization}} \\
Optimizer & AdamW \\
Learning rate & $2 \times 10^{-5}$ \\
LR schedule & cosine, 5\% linear warmup \\
Gradient clipping & max norm $1.0$ \\
$\beta_1, \beta_2$ & $0.9, 0.95$ \\
Batch size (effective) & $8$ ($4$/GPU $\times$ $2$ GPUs) \\
Epochs & $1$ \\
Max sequence length & $1024$ \\
Precision & bfloat16 \\
\midrule
\multicolumn{2}{@{}l}{\textit{Unrolling \& Loss}} \\
$K$ (unroll steps) & $4$ \\
$\lambda_d$ (dense reveal mix) & $0.75$ \\
$\lambda_s$ (state norm weight) & $10^{-4}$ \\
$\tau$ (state norm threshold) & $1.0$ \\
\midrule
\multicolumn{2}{@{}l}{\textit{Inference}} \\
Max generation length & $256$ \\
Block size & $32$ \\
KV cache mode & dual \\
Confidence threshold & $0.9$ \\
\bottomrule
\vspace{-2em}
\end{tabular}
\end{table}

\subsection{Model Parameter Analysis}
\label{sec:param-analysis}

All recurrent modules operate in bottleneck dimensions $d_m{=}d_b{=}768$ and memory slot dimension $D_s{=}1024$ rather than the backbone hidden size $D$, which keeps the parameter count compact.
Only three interface projections, the Mixer's backbone down-projection $\mathbf{W}^{h}_{\downarrow}\!\in\!\mathbb{R}^{d_m \times D}$, and the Injector's input/output projections $\mathbf{W}^{e}_{\downarrow}\!\in\!\mathbb{R}^{d_b \times D}$, $\mathbf{W}_{\uparrow}\!\in\!\mathbb{R}^{D \times d_b}$, depend on the backbone hidden size $D$. All other parameters are shared the same across backbones.
Table~\ref{tab:param-breakdown} provides a per-module breakdown.

\begin{table}[ht]
\centering
\small
\vspace{-1.8em}
\caption{Per-module parameter breakdown of the MetaState recurrent modules for each backbone ($D_s{=}1024$, $d_m{=}d_b{=}768$, $M{=}64$). Rows marked $^{\dagger}$ are the only backbone-dependent components.}
\label{tab:param-breakdown}
\begin{tabular}{@{}lrr@{}}
\toprule
Module / Component & LLaDA & Dream \\
\midrule
\textsc{SharedTimeConditioner} & 2,102,016 & 2,102,016 \\
\quad Sinusoidal MLP ($256{\to}1024{\to}1024$) & 1,312,768 & 1,312,768 \\
\quad Content path (norm + proj + gate) & 789,248 & 789,248 \\
\midrule
\textsc{Mixer} & 17,899,264 & 17,506,048 \\
\quad Slot self-attention ($D_s$ space, GQA) & 3,146,752 & 3,146,752 \\
\quad Down-projections$^{\dagger}$ ($d_m{\times}D_s + d_m{\times}D$) & 3,932,160 & 3,538,944 \\
\quad Cross-attention + AdaRMSNorms & 6,494,976 & 6,494,976 \\
\quad SwiGLU FFN + up-projection & 4,325,376 & 4,325,376 \\
\midrule
\textsc{Updater} & 10,626,048 & 10,626,048 \\
\quad AdaRMSNorms (${\times}2$) & 4,200,448 & 4,200,448 \\
\quad GRU projections (gate + candidate) & 6,294,528 & 6,294,528 \\
\quad Learnable $\mathbf{s}_0$ + slot embeddings & 131,072 & 131,072 \\
\midrule
\textsc{Injector} & 20,457,217 & 19,670,785 \\
\quad Interface projections$^{\dagger}$ ($d_b{\times}D + D{\times}d_b$) & 6,291,456 & 5,505,024 \\
\quad Self-attention (RMSNorm, GQA) & 1,770,241 & 1,770,241 \\
\quad State proj + cross-attn + AdaRMSNorms & 8,856,576 & 8,856,576 \\
\quad SwiGLU FFN & 3,538,944 & 3,538,944 \\
\midrule
\textbf{Total trainable} & \textbf{51,084,545} & \textbf{49,904,897} \\
Frozen backbone & 8,056,602,369 & 7,655,458,049 \\
\textbf{Overhead} & \textbf{0.63\%} & \textbf{0.65\%} \\
\bottomrule
\end{tabular}
\vspace{-1em}
\end{table}

\subsection{Soft Diffusion Hyperparameter Details}
\label{sec:soft-diffusion-details}

Soft Diffusion, introduced in Latent Refinement Decoding (LRD)~\citep{zhu2025latent}, replaces the hard remasking at mask positions with a weighted mixture of token embeddings. For each mask position $i$ with sampled token $\hat{x}_i$, the input embedding becomes:
\begin{equation*}
\tilde{\mathbf{e}}_i = (1 - r_f) \cdot \mathbf{e}_{\hat{x}_i}
  + r_f \cdot \textstyle\sum\nolimits_{v} p_v \cdot \mathbf{e}_v,
\end{equation*}
where $p_v$ is the model's predicted probability for token $v$ (after nucleus filtering with threshold $p$), $r_f \in [0,1]$ is the mix ratio factor controlling the mixture weight, and $\mathbf{e}_v$ denotes the token embedding for vocabulary entry $v$. The two key hyperparameters are the nucleus probability threshold (top-$p$) and the mix ratio factor ($r_f$).

Following the practice of LRD~\citep{zhu2025latent}, we consider top-$p \in \{0.2, 0.9\}$, since the paper reports that values $\geq 0.2$ are effective, and $r_f \in \{0.1, 0.15, 0.2\}$, which spans the effective range $[0.1, 0.2]$ reported in the paper. For each method--backbone--benchmark combination, we evaluate all six $(p, r_f)$ pairs listed above and report the best-performing configuration. Table~\ref{tab:soft-diffusion-params} lists the selected hyperparameters.

\begin{table}[ht]
\centering
\small
\vspace{-1em}
\caption{Best Soft Diffusion hyperparameters $(p, r_f)$ for each method, backbone, and benchmark. All values selected by grid search over $p \in \{0.2, 0.9\}$ and $r_f \in \{0.1, 0.15, 0.2\}$.}
\label{tab:soft-diffusion-params}
\begin{tabular}{@{}ll cccc@{}}
\toprule
Backbone & Method & GSM8K & MATH & HE & MBPP \\
\midrule
Dream-Instruct & Soft Diffusion & (0.9,\,0.15) & (0.9,\,0.1) & (0.9,\,0.2) & (0.9,\,0.2) \\
Dream-Instruct & \quad +\,MetaState & (0.2,\,0.1) & (0.2,\,0.2) & (0.9,\,0.1) & (0.2,\,0.1) \\
\midrule
LLaDA-Instruct & Soft Diffusion & (0.2,\,0.1) & (0.2,\,0.15) & (0.9,\,0.2) & (0.2,\,0.2) \\
LLaDA-Instruct & \quad +\,MetaState & (0.2,\,0.1) & (0.9,\,0.2) & (0.2,\,0.15) & (0.2,\,0.2) \\
\bottomrule
\end{tabular}
\end{table}

To check that the compatibility conclusion is not driven only by the best selected hyperparameter in each cell, we also average over all six Soft Diffusion grid configurations. Table~\ref{tab:soft-diffusion-average} shows that Soft Diffusion combined with MetaState improves six of eight backbone--benchmark averages, with only small drops on LLaDA-Instruct MATH-500 and Dream-Instruct HumanEval. This supports the same conclusion as the best-configuration table: Soft Diffusion and MetaState are largely complementary, while occasional small drops can occur from the embedding-mixture input distribution shift.

\begin{table}[ht]
\centering
\small
\vspace{-1em}
\caption{Soft Diffusion average over all six grid configurations. Averaging removes dependence on a single selected $(p,r_f)$ pair.}
\label{tab:soft-diffusion-average}
\begin{tabular}{@{}l l c c c@{}}
\toprule
Backbone & Metric & Soft Diff. avg & Soft Diff. + Meta avg & $\Delta$ \\
\midrule
LLaDA-Instruct & GSM8K flexible EM & 78.14 & 79.48 & +1.34 \\
LLaDA-Instruct & HumanEval pass@1 & 36.89 & 39.12 & +2.23 \\
LLaDA-Instruct & MBPP pass@1 & 28.27 & 31.77 & +3.50 \\
LLaDA-Instruct & MATH-500 math verify & 37.23 & 37.20 & -0.03 \\
Dream-Instruct & GSM8K flexible EM & 76.25 & 78.27 & +2.02 \\
Dream-Instruct & HumanEval pass@1 & 57.52 & 56.91 & -0.61 \\
Dream-Instruct & MBPP pass@1 & 51.24 & 54.28 & +3.04 \\
Dream-Instruct & MATH-500 math verify & 44.20 & 44.83 & +0.63 \\
\bottomrule
\end{tabular}
\vspace{-1em}
\end{table}

\subsection{Comparison with LoRA Fine-Tuning}
\label{sec:lora-comparison}

MetaState and LoRA~\citep{hu2022lora} both introduce a small number of trainable parameters on top of a frozen backbone, but they operate at fundamentally different levels. LoRA injects low-rank updates into the backbone's linear layers, directly modifying the model's internal representations and thereby its learned inner capabilities. MetaState, by contrast, leaves all backbone weights unchanged and operates entirely at the denoising interface: it reads from the backbone hidden states after each forward pass through the Mixer and writes a lightweight additive signal into the token embeddings before the next pass through the Injector. No gradient flows into the backbone during MetaState training, and the backbone's per-step predictive behavior is affected only through this external recurrent state channel. In this sense, MetaState introduces cross-step information passing without enhancing the model's internal capability, whereas LoRA improves single-step capability.

Because the two methods affect different aspects of the generation process, this comparison should not be interpreted as a head-to-head evaluation of fine-tuning methods. Nevertheless, a LoRA baseline under a matched training setup helps rule out an alternative explanation of MetaState's gains: namely, that the improvements arise primarily from exposure to the T\"{u}lu-3 training distribution rather than from the working-memory mechanism itself. To test this possibility, we fine-tune the same Instruct backbones with LoRA ($r{=}32$, $\alpha{=}64$, \texttt{target\_modules\,=\,all-linear}) on the identical 50k T\"{u}lu-3 sequences, using the same optimizer, learning-rate schedule, and training budget described in Appendix~\ref{sec:experimental-details}.

\begin{table}[ht]
\vspace{-1em}
\caption{Comparison with LoRA fine-tuning on Instruct backbones. LoRA ($r{=}32$, $\alpha{=}64$, all-linear) is trained on the same 50k T\"{u}lu-3 data with matched optimization hyperparameters. $\dagger$\,denotes\,`+\,Soft Diffusion'. \textbf{Bold} marks the best result per column.}
\label{tab:lora-comparison}
\centering
\small
\begin{tabular}{@{}l cc @{\hspace{16pt}} cc @{\hspace{16pt}} c@{}}
\toprule
\toprule
& \multicolumn{2}{c}{\textit{Dream-Instruct\ (7B)}}
& \multicolumn{2}{c}{\textit{LLaDA-Instruct\ (8B)}} & \\
\cmidrule(lr){2-3} \cmidrule(lr){4-5}
Method & GSM8K & HumanEval & GSM8K & HumanEval & \textbf{Avg.} \\
\midrule
Instruct
  & 74.8 & 56.1
  & 78.5 & 37.2 & \cellcolor{orange!8}61.7 \\
\quad + LoRA
  & 76.0 & 55.5
  & 79.2 & 39.0 & \cellcolor{orange!8}62.4 \\
\midrule
\quad + MetaState
  & 78.1 & \textbf{59.8}
  & 79.5 & 39.6 & \cellcolor{orange!8}64.3 \\
\quad + MetaState$^\dagger$
  & \textbf{79.4} & 59.2
  & \textbf{80.3} & \textbf{40.9} & \cellcolor{orange!8}\textbf{65.0} \\
\bottomrule
\bottomrule
\end{tabular}
\end{table}

Table~\ref{tab:lora-comparison} shows that LoRA underperforms MetaState on every benchmark pair. On Dream-Instruct, LoRA even slightly reduces HumanEval accuracy relative to the original backbone ($55.5$ vs.\ $56.1$), whereas MetaState improves it to $59.8$. On LLaDA-Instruct, LoRA narrows the gap but still remains below MetaState on both GSM8K ($79.2$ vs.\ $79.5$) and HumanEval ($39.0$ vs.\ $39.6$). When MetaState is further combined with Soft Diffusion (last row), the margin widens on all benchmarks. Since LoRA directly updates backbone weights and therefore has greater capacity to absorb distributional regularities from the training data than MetaState's frozen-backbone design, its inferior performance suggests that MetaState's gains cannot be explained simply by data-domain exposure. Instead, the improvements should be attributed to the persistent working memory introduced by MetaState: by carrying continuous information across denoising steps, MetaState alleviates the Information Island issue (\S\ref{sec:info-island}) in a way that LoRA's single-step weight adaptation cannot replicate.

\subsection{Robustness Across Decoding Configurations}
\label{sec:decoding-robustness}

The main results use generation length 256, block size 32, and confidence threshold 0.9. We add three sweeps to test whether MetaState's gains depend on this single decoding configuration.

First, Table~\ref{tab:rebuttal-block-size} varies block size, which directly changes the amount of information that must be preserved within each per-block denoising trajectory. MetaState improves all tested block sizes on both Dream-Instruct and LLaDA-Instruct. The gains become especially large for Dream at block sizes 64 and 128, where the original backbone degrades sharply and MetaState preserves much more accuracy. We do not interpret block size 256 as part of a monotonic trend, because the reveal process becomes aggressive and both the original and augmented models degrade substantially.

\begin{table}[t]
\centering
\small
\vspace{-1em}
\caption{GSM8K block-size sweep. Block size changes the per-block denoising trajectory length; MetaState is positive at all tested sizes.}
\label{tab:rebuttal-block-size}
\begin{tabular}{@{}l c c c c@{}}
\toprule
Backbone & Block size & Original & MetaState & Gain \\
\midrule
Dream-Instruct & 16 & 78.24 & 78.85 & +0.61 \\
Dream-Instruct & 32 & 74.75 & 78.09 & +3.34 \\
Dream-Instruct & 64 & 65.43 & 77.18 & +11.75 \\
Dream-Instruct & 128 & 52.01 & 74.83 & +22.82 \\
Dream-Instruct & 256 & 50.34 & 53.07 & +2.73 \\
LLaDA-Instruct & 16 & 77.94 & 78.62 & +0.68 \\
LLaDA-Instruct & 32 & 78.47 & 79.53 & +1.06 \\
LLaDA-Instruct & 64 & 76.88 & 78.01 & +1.13 \\
LLaDA-Instruct & 128 & 71.95 & 75.66 & +3.71 \\
LLaDA-Instruct & 256 & 52.46 & 53.45 & +0.99 \\
\bottomrule

\end{tabular}
\vspace{-1em}
\end{table}

Second, Table~\ref{tab:rebuttal-threshold-sweep} varies the confidence threshold under the main generation length and block size. Lower thresholds reduce the effective number of denoising updates. MetaState remains positive in all 28 matched GSM8K cells across thresholds from 0.60 to 0.95, indicating that the gain is not restricted to a single slow or fast decoding budget.

\begin{table*}[t]
\centering
\small
\vspace{-1em}
\caption{GSM8K confidence-threshold sweep at generation length 256, block size 32, and dual cache. Lower thresholds reduce the effective number of denoising updates. MetaState improves all matched cells.}
\label{tab:rebuttal-threshold-sweep}
\begin{tabular}{@{}l c c c c c c@{}}
\toprule
Model & Threshold & Original & MetaState & Gain & Orig. fwd/seq & Meta fwd/seq \\
\midrule
Dream-Base & 0.95 & 73.16 & 76.42 & +3.26 & 156.34 & 78.50 \\
Dream-Base & 0.85 & 70.43 & 75.28 & +4.85 & 139.10 & 67.14 \\
Dream-Base & 0.80 & 70.58 & 74.15 & +3.57 & 131.05 & 63.10 \\
Dream-Base & 0.75 & 69.60 & 73.16 & +3.56 & 123.63 & 58.82 \\
Dream-Base & 0.70 & 66.26 & 70.20 & +3.94 & 117.02 & 54.75 \\
Dream-Base & 0.65 & 64.37 & 67.48 & +3.11 & 110.39 & 51.78 \\
Dream-Base & 0.60 & 60.27 & 64.67 & +4.40 & 103.88 & 48.74 \\
Dream-Instruct & 0.95 & 75.59 & 79.76 & +4.17 & 69.60 & 74.47 \\
Dream-Instruct & 0.85 & 73.84 & 77.94 & +4.10 & 59.93 & 65.41 \\
Dream-Instruct & 0.80 & 74.15 & 77.33 & +3.18 & 56.37 & 61.84 \\
Dream-Instruct & 0.75 & 71.65 & 73.77 & +2.12 & 52.46 & 55.64 \\
Dream-Instruct & 0.70 & 68.23 & 72.48 & +4.25 & 49.12 & 52.67 \\
Dream-Instruct & 0.65 & 65.88 & 68.46 & +2.58 & 45.91 & 50.83 \\
Dream-Instruct & 0.60 & 63.61 & 63.91 & +0.30 & 43.27 & 47.11 \\
LLaDA-Base & 0.95 & 68.46 & 78.09 & +9.63 & 153.58 & 100.40 \\
LLaDA-Base & 0.85 & 68.16 & 79.23 & +11.07 & 128.34 & 74.47 \\
LLaDA-Base & 0.80 & 68.16 & 77.71 & +9.55 & 119.36 & 66.72 \\
LLaDA-Base & 0.75 & 68.01 & 77.41 & +9.40 & 111.92 & 59.64 \\
LLaDA-Base & 0.70 & 67.17 & 77.63 & +10.46 & 105.04 & 54.04 \\
LLaDA-Base & 0.65 & 66.11 & 76.50 & +10.39 & 98.16 & 48.93 \\
LLaDA-Base & 0.60 & 61.71 & 72.93 & +11.22 & 91.93 & 44.35 \\
LLaDA-Instruct & 0.95 & 76.80 & 78.47 & +1.67 & 101.01 & 105.67 \\
LLaDA-Instruct & 0.85 & 75.97 & 78.32 & +2.35 & 75.03 & 79.42 \\
LLaDA-Instruct & 0.80 & 75.51 & 77.26 & +1.75 & 66.72 & 71.34 \\
LLaDA-Instruct & 0.75 & 76.35 & 76.57 & +0.22 & 59.68 & 63.86 \\
LLaDA-Instruct & 0.70 & 74.30 & 77.03 & +2.73 & 54.09 & 57.62 \\
LLaDA-Instruct & 0.65 & 75.44 & 75.83 & +0.39 & 48.87 & 52.36 \\
LLaDA-Instruct & 0.60 & 71.87 & 72.78 & +0.91 & 44.56 & 47.64 \\
\bottomrule
\end{tabular}
\end{table*}

Third, Table~\ref{tab:rebuttal-generation-length} varies the maximum generation length on Dream-Instruct. MetaState improves both GSM8K and MBPP at all tested lengths from 128 to 1024 tokens, showing that the effect is not specific to the default 256-token setting.

\begin{table}[t]
\centering
\small
\vspace{-1em}
\caption{Dream-Instruct generation-length sweep. MetaState remains positive on GSM8K and MBPP from 128 to 1024 tokens.}
\label{tab:rebuttal-generation-length}
\begin{tabular}{@{}c c c c c c c@{}}
\toprule
Length & GSM8K Orig. & GSM8K Meta & Gain & MBPP Orig. & MBPP Meta & Gain \\
\midrule
128 & 69.98 & 71.87 & +1.89 & 49.80 & 52.60 & +2.80 \\
256 & 74.75 & 78.09 & +3.34 & 51.80 & 55.00 & +3.20 \\
384 & 75.89 & 78.85 & +2.96 & 50.20 & 55.00 & +4.80 \\
512 & 75.59 & 79.83 & +4.24 & 50.40 & 54.20 & +3.80 \\
768 & 74.30 & 79.53 & +5.23 & 51.40 & 55.20 & +3.80 \\
1024 & 75.51 & 78.70 & +3.19 & 52.20 & 54.60 & +2.40 \\
\bottomrule
\end{tabular}
\vspace{-1em}
\end{table}

Finally, Table~\ref{tab:rebuttal-multiseed} reports a GSM8K multi-seed check. MetaState improves all four settings across seeds, with large gains on Base backbones and smaller positive improvements on Instruct backbones.

\begin{table}[t]
\centering
\small
\caption{GSM8K multi-seed robustness. MetaState improves all four settings across seeds, with large gains on Base backbones and smaller improvements on Instruct backbones.}
\label{tab:rebuttal-multiseed}
\begin{tabular}{@{}l c c c c@{}}
\toprule
Model & Original & MetaState & Gain & Gain / pooled std \\
\midrule
Dream-Base & $73.92 \pm 0.80$ & $76.00 \pm 0.88$ & +2.08 & 2.47 \\
Dream-Instruct & $75.19 \pm 0.88$ & $78.64 \pm 0.67$ & +3.45 & 4.41 \\
LLaDA-Base & $67.87 \pm 0.75$ & $77.67 \pm 0.67$ & +9.80 & 13.78 \\
LLaDA-Instruct & $78.16 \pm 0.86$ & $78.92 \pm 0.98$ & +0.76 & 0.82 \\
\bottomrule
\end{tabular}
\end{table}

\subsection{Hyperparameter Sensitivity}
\label{sec:hyperparam-sensitivity}

\begin{figure}
    \centering
    \includegraphics[width=\linewidth]{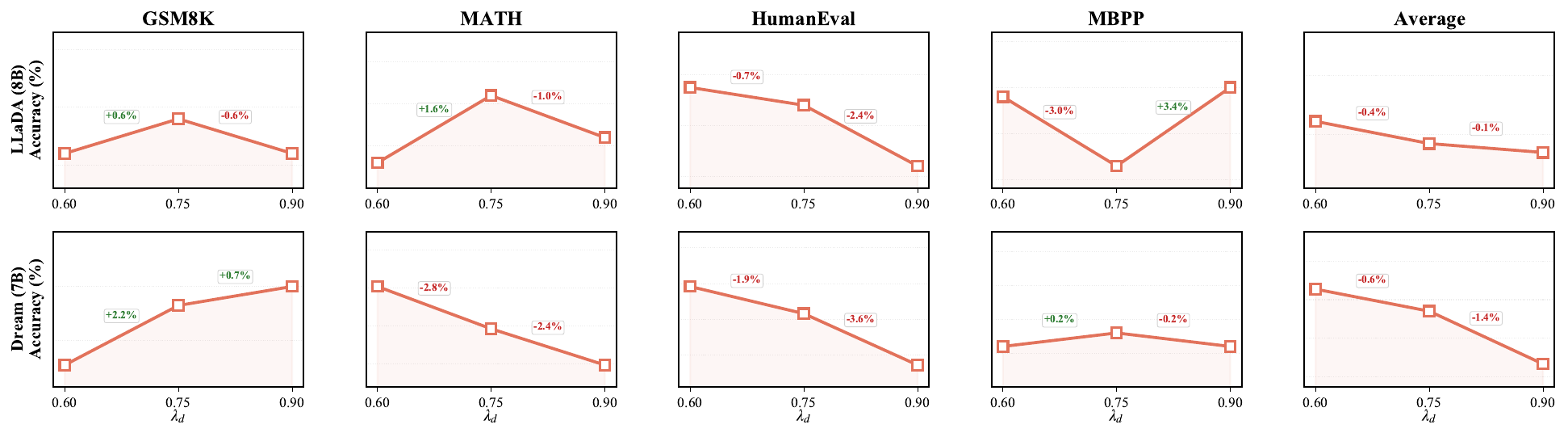}
    \vspace{-1.5em}
    \caption{Sensitivity to the dense-reveal loss mixing ratio $\lambda_d$. Increasing $\lambda_d$ from 0.6 to 0.9 yields modest drops in average scores.}
    \label{fig:ablation-drm}
    \vspace{-1.5em}
\end{figure}

\begin{figure}[ht]
    \centering
    \includegraphics[width=\linewidth]{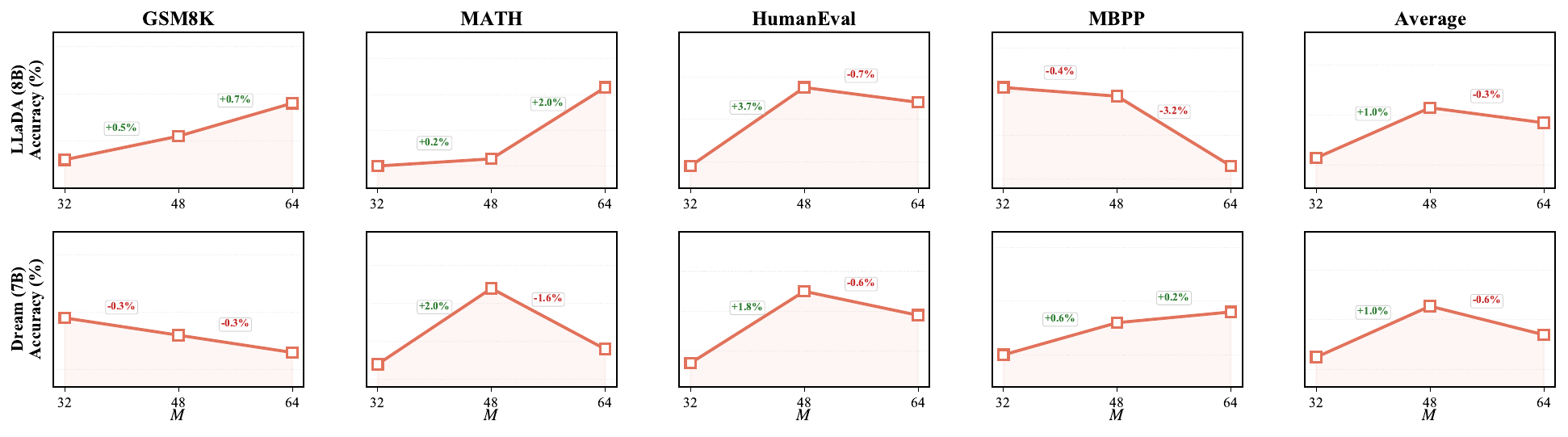}
    \vspace{-1.5em}
    \caption{Sensitivity to the number of memory slots $M$. Increasing $M$ from 32 to 48 yields modest gains on most benchmarks, while further increasing to 64 produces modest drops.}
    \label{fig:ablation-nms}
    \vspace{-0.5em}
\end{figure}

\begin{figure}[ht]
    \centering
    \includegraphics[width=\linewidth]{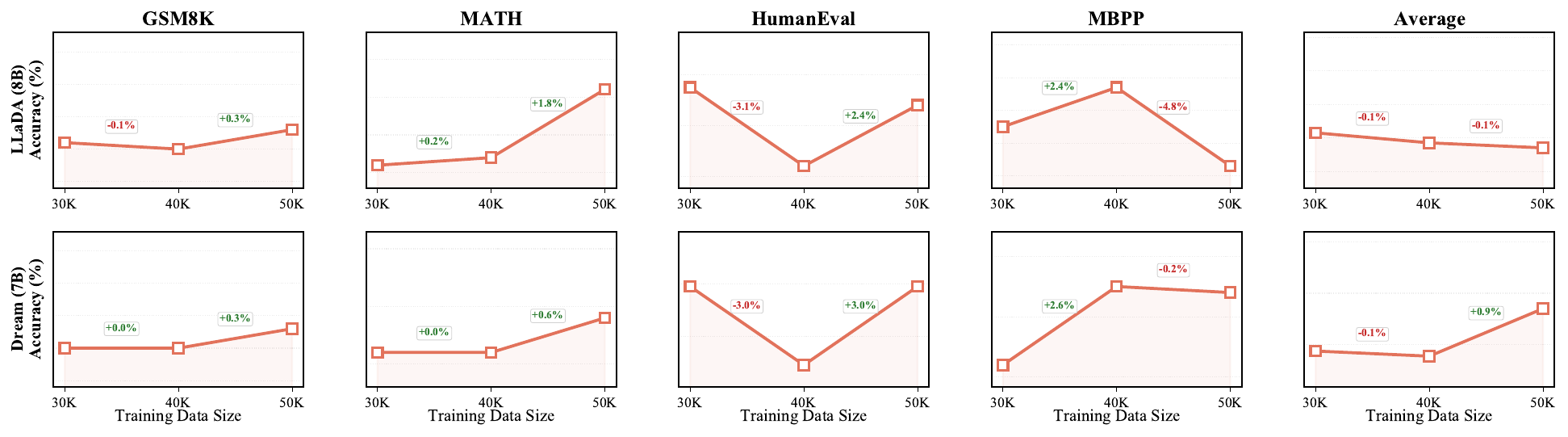}
    \vspace{-1.5em}
    \caption{Sensitivity to training data size. MetaState achieves competitive performance even with 30k training examples, and accuracy remains stable as the dataset increases to 50k.}
    \label{fig:ablation-train-size}
    \vspace{-0.5em}
\end{figure}

\begin{figure}[ht]
    \centering
    \includegraphics[width=\linewidth]{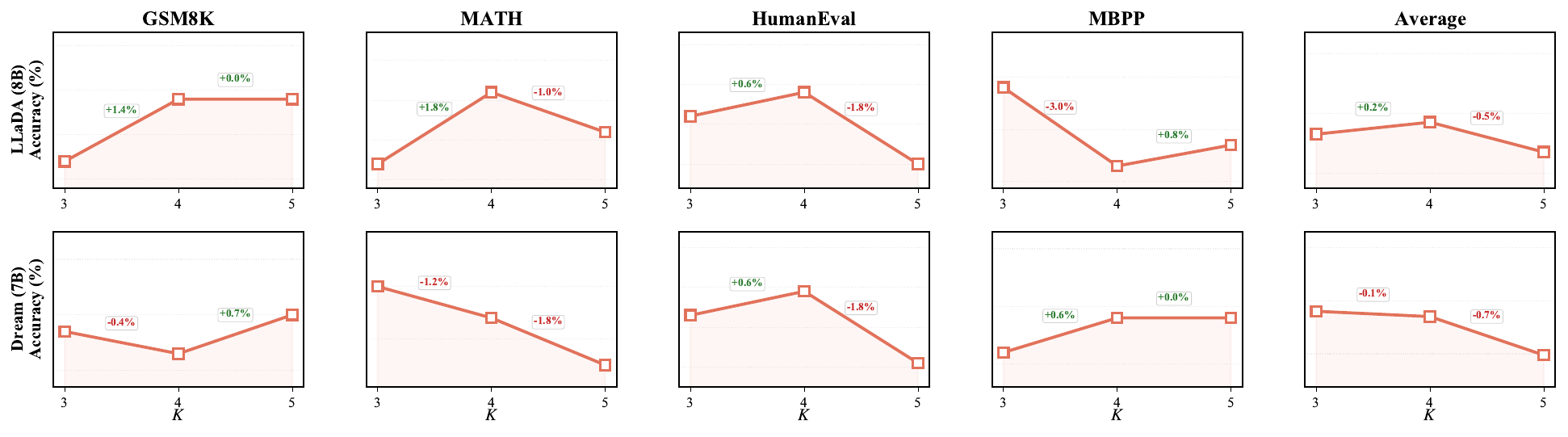}
    \vspace{-1.5em}
    \caption{Sensitivity to the unroll depth $K$. Performance is broadly stable across $K \in \{3,4,5\}$, with average accuracy varying by less than one point.}
    \label{fig:ablation-us}
    \vspace{-0.5em}
\end{figure}

To evaluate the robustness of MetaState, we vary four hyperparameters individually while keeping all others fixed at their default values in Table~\ref{tab:training-hyperparams}. Specifically, we sweep the dense-reveal loss mixing ratio $\lambda_d \in \{0.60, 0.75, 0.90\}$ (default: $0.75$), the number of memory slots $M \in \{32, 48, 64\}$ (default: $64$), the training set size $\in \{30\text{k}, 40\text{k}, 50\text{k}\}$ (default: $50\text{k}$), and the unroll depth $K \in \{3, 4, 5\}$ (default: $4$). All experiments use the same optimizer, learning-rate schedule, and evaluation protocol described in Appendix~\ref{sec:experimental-details}, and are conducted on two NVIDIA A100 GPUs. Figures~\ref{fig:ablation-drm}--\ref{fig:ablation-us} report per-task accuracy on both Dream-Instruct and LLaDA-Instruct backbones.

Across all four sweeps, performance varies only within a relatively narrow range on both backbones and across all benchmarks. Overall, these results suggest that MetaState is not overly sensitive to a single hyperparameter, and that the default setting represents a robust operating choice rather than a narrowly tuned optimum.

\FloatBarrier

\subsection{Case Study}
\label{sec:case-study}

To provide a qualitative view of how MetaState influences the denoising process, we present case studies that compare the full denoising trajectories of the original LLaDA-Instruct and Dream-Instruct backbones with their counterparts augmented by MetaState on mathematical reasoning examples. All experiments in this section are conducted on NVIDIA A100 GPUs. The results show a clear qualitative pattern: with MetaState, the denoising trajectory is more likely to preserve correct intermediate computations, organize the correct solution structure earlier, and maintain a coherent reasoning path through later steps. These observations provide evidence that MetaState mitigates the Information Island issue by carrying forward useful intermediate information that would otherwise be lost across the discrete sampling-and-remasking interface.
\vspace{1em}

\begin{tcolorbox}[title={Case Study 1: \textit{LLaDA-Instruct (8B)} vs. \textit{LLaDA-Instruct (8B) + MetaState}}, fontupper=\small, fonttitle=\small,
  boxrule=0.4pt,
  arc=1mm,
  left=1mm,
  right=1mm,
  top=0.6mm,
  bottom=0.6mm,
  boxsep=0.5mm,
  before skip=0mm,
  after skip=0mm,
  halign upper=left,
  valign=top,
  breakable
]
\textbf{Question:}
\\
Toula went to the bakery and bought various types of pastries. She bought 3 dozen donuts which cost \$68 per dozen, 2 dozen mini cupcakes which cost \$80 per dozen, and 6 dozen mini cheesecakes for \$55 per dozen. How much was the total cost?

Please solve this step by step and end with \#\#\#\# \textless answer\textgreater.
\\

\textbf{Ground-truth answer:}
\(\mathbf{694}\)

\noindent\rule{\linewidth}{0.5pt}

\medskip
\textbf{Denoising Timestep \(t = 0.80\)}

\begin{tcbraster}[raster columns=2, raster equal height=rows, raster column skip=2mm]
\begin{tcolorbox}[colback=red!3,colframe=red!35!black,boxrule=0.4pt,arc=1mm,title={\textit{LLaDA-Instruct (8B)}}]
{\ttfamily
Cost of donuts:\masktok{MASKx1}3\masktok{MASKx90}
}
\end{tcolorbox}
\begin{tcolorbox}[colback=blue!3,colframe=blue!35!black,boxrule=0.4pt,arc=1mm,title={\textit{LLaDA-Instruct (8B) + MetaState}}]
{\ttfamily
The cost of the donuts is\masktok{MASKx90}
}
\end{tcolorbox}
\end{tcbraster}

\medskip
\textbf{Denoising Timestep \(t = 0.60\)}

\begin{tcbraster}[raster columns=2, raster equal height=rows, raster column skip=2mm]
\begin{tcolorbox}[colback=red!3,colframe=red!35!black,boxrule=0.4pt,arc=1mm]
{\ttfamily
Cost of donuts: 3 dozen\masktok{MASKx1} \$68\masktok{MASKx1} dozen\masktok{MASKx8} mini cupcakes: 2 dozen\masktok{MASKx1} \$80\masktok{MASKx65}
}
\end{tcolorbox}
\begin{tcolorbox}[colback=blue!3,colframe=blue!35!black,boxrule=0.4pt,arc=1mm]
{\ttfamily
The cost of the donuts is 3\masktok{MASKx1} \$68 = \$204\masktok{MASKx25} mini\masktok{MASKx53}
}
\end{tcolorbox}
\end{tcbraster}

\medskip
\textbf{Denoising Timestep \(t = 0.40\)}

\begin{tcbraster}[raster columns=2, raster equal height=rows, raster column skip=2mm]
\begin{tcolorbox}[colback=red!3,colframe=red!35!black,boxrule=0.4pt,arc=1mm]
{\ttfamily
Cost of donuts: 3 dozen\masktok{MASKx1} \$68 per dozen = \$204\par
Cost of mini cupcakes: 2 dozen\masktok{MASKx1} \$80 per dozen = \$160\par
Cost of mini cheesecakes: 6 dozen * \$55 per dozen = \$330\par
Total cost:\masktok{MASKx32}
}
\end{tcolorbox}
\begin{tcolorbox}[colback=blue!3,colframe=blue!35!black,boxrule=0.4pt,arc=1mm]
{\ttfamily
The cost of the donuts is 3\masktok{MASKx1} \$68 = \$204\masktok{MASKx15} \$160.\masktok{MASKx1}The cost of the mini cheesecakes is 6\masktok{MASKx1} \$55 = \$330.\par
\masktok{MASKx36}
}
\end{tcolorbox}
\end{tcbraster}

\medskip
\textbf{Denoising Timestep \(t = 0.20\)}

\begin{tcbraster}[raster columns=2, raster equal height=rows, raster column skip=2mm]
\begin{tcolorbox}[colback=red!3,colframe=red!35!black,boxrule=0.4pt,arc=1mm]
{\ttfamily
Cost of donuts: 3 dozen * \$68 per dozen = \$204\par
Cost of mini cupcakes: 2 dozen * \$80 per dozen = \$160\par
Cost of mini cheesecakes: 6 dozen * \$55 per dozen = \$330\par
Total cost: \$204 + \$160 + \$330 = \colorbox{red!15}{\$794}\par
\#\#\#\#\masktok{MASKx6}9\masktok{MASKx1}
}
\end{tcolorbox}
\begin{tcolorbox}[colback=blue!3,colframe=blue!35!black,boxrule=0.4pt,arc=1mm]
{\ttfamily
The cost of the donuts is 3 * \$68 = \$204\masktok{MASKx6} cupcakes is 2 * \$80 = \$160.\par
The cost of the mini cheesecakes is 6 * \$55 = \$330.\par
\masktok{MASKx31} \masktok{MASKx1}94
}
\end{tcolorbox}
\end{tcbraster}

\medskip
\textbf{Denoising Timestep \(t = 0.00\)}

\begin{tcbraster}[raster columns=2, raster equal height=rows, raster column skip=2mm]
\begin{tcolorbox}[colback=red!3,colframe=red!35!black,boxrule=0.4pt,arc=1mm]
{\ttfamily
Cost of donuts: 3 dozen * \$68 per dozen = \$204\par
Cost of mini cupcakes: 2 dozen * \$80 per dozen = \$160\par
Cost of mini cheesecakes: 6 dozen * \$55 per dozen = \$330\par
Total cost: \$204 + \$160 + \$330 = \colorbox{red!15}{\$794}\par
\#\#\#\#
\colorbox{red!15}{794}
}
\end{tcolorbox}
\begin{tcolorbox}[colback=blue!3,colframe=blue!35!black,boxrule=0.4pt,arc=1mm]
{\ttfamily
The cost of the donuts is 3 * \$68 = \$204 cupcakes is 2 * \$80 = \$160.\par
The cost of the mini cheesecakes is 6 * \$55 = \$330.\par
 cost is \$204 + \$160 + \$330 = \colorbox{green!15}{\$694}.\par
\#\#\#\# \colorbox{green!15}{694}
}
\end{tcolorbox}
\end{tcbraster}

\end{tcolorbox}

\vspace{0.8em}

\vspace{0.8em}
\begin{tcolorbox}[title={Case Study 2: \textit{Dream-Instruct (7B)} vs. \textit{Dream-Instruct (7B) + MetaState}}, fontupper=\small, fonttitle=\small,
  boxrule=0.4pt,
  arc=1mm,
  left=1mm,
  right=1mm,
  top=0.6mm,
  bottom=0.6mm,
  boxsep=0.5mm,
  before skip=0mm,
  after skip=0mm,
  halign upper=left,
  valign=top,
  breakable
]
\textbf{Question:}
\\
Gloria is shoe shopping when she comes across a pair of boots that fit her shoe budget. However, she has to choose between the boots and two pairs of high heels that together cost five dollars less than the boots. If one pair of heels costs \$33 and the other costs twice as much, how many dollars are the boots?
\\
Please solve this step by step and end with \#\#\#\# \textless answer\textgreater.
\\

\textbf{Ground-truth answer:}
\(\mathbf{104}\)

\noindent\rule{\linewidth}{0.5pt}

\medskip
\textbf{Denoising Timestep \(t = 0.80\)}

\begin{tcbraster}[raster columns=2, raster equal height=rows, raster column skip=2mm]
\begin{tcolorbox}[colback=red!3,colframe=red!35!black,boxrule=0.4pt,arc=1mm,title={\textit{Dream-Instruct (7B)}}]
{\ttfamily
\masktok{MASKx80}.\par
\#\#\#\# 13\masktok{MASKx1}\par
The answer is: 1\masktok{MASKx2}
}
\end{tcolorbox}
\begin{tcolorbox}[colback=blue!3,colframe=blue!35!black,boxrule=0.4pt,arc=1mm,title={\textit{Dream-Instruct (7B) + MetaState}}]
{\ttfamily
If one pair of heels costs \$33\masktok{MASKx1} the\masktok{MASKx83}
}
\end{tcolorbox}
\end{tcbraster}

\medskip
\textbf{Denoising Timestep \(t = 0.60\)}

\begin{tcbraster}[raster columns=2, raster equal height=rows, raster column skip=2mm]
\begin{tcolorbox}[colback=red!3,colframe=red!35!black,boxrule=0.4pt,arc=1mm]
{\ttfamily
\masktok{MASKx65} the boots\masktok{MASKx1} \$13\masktok{MASKx1} + \$5 = \$13\masktok{MASKx1}.\par
\#\#\#\# 13\masktok{MASKx1}\par
The answer is: 13\masktok{MASKx1}
}
\end{tcolorbox}
\begin{tcolorbox}[colback=blue!3,colframe=blue!35!black,boxrule=0.4pt,arc=1mm]
{\ttfamily
If one pair of heels costs \$33 and the other costs twice as much, then the\masktok{MASKx1} pair of heels costs\masktok{MASKx6} =\masktok{MASKx1}66\masktok{MASKx58}.\masktok{MASKx1}
}
\end{tcolorbox}
\end{tcbraster}

\medskip
\textbf{Denoising Timestep \(t = 0.40\)}

\begin{tcbraster}[raster columns=2, raster equal height=rows, raster column skip=2mm]
\begin{tcolorbox}[colback=red!3,colframe=red!35!black,boxrule=0.4pt,arc=1mm]
{\ttfamily
\masktok{MASKx3} of\masktok{MASKx52} cost five dollars less than the boots,\masktok{MASKx1} the boots cost \$132 + \$5 = \colorbox{red!15}{\$137}.\par
\#\#\#\# \colorbox{red!15}{137}\par
The answer is: \colorbox{red!15}{137}
}
\end{tcolorbox}
\begin{tcolorbox}[colback=blue!3,colframe=blue!35!black,boxrule=0.4pt,arc=1mm]
{\ttfamily
If one pair of heels costs \$33 and the other costs twice as much, then the second pair of heels costs \$33\masktok{MASKx1} 2 = \$66.\par
\masktok{MASKx53} \colorbox{green!15}{104}.\masktok{MASKx1}
}
\end{tcolorbox}
\end{tcbraster}

\medskip
\textbf{Denoising Timestep \(t = 0.20\)}

\begin{tcbraster}[raster columns=2, raster equal height=rows, raster column skip=2mm]
\begin{tcolorbox}[colback=red!3,colframe=red!35!black,boxrule=0.4pt,arc=1mm]
{\ttfamily
If one pair of heels costs \$33, then the other\masktok{MASKx41} heels together cost five dollars less than the boots, then the boots cost \$132 + \$5 = \colorbox{red!15}{\$137}.\par
\#\#\#\# \colorbox{red!15}{137}\par
The answer is: \colorbox{red!15}{137}
}
\end{tcolorbox}
\begin{tcolorbox}[colback=blue!3,colframe=blue!35!black,boxrule=0.4pt,arc=1mm]
{\ttfamily
If one pair of heels costs \$33 and the other costs twice as much, then the second pair of heels costs \$33\masktok{MASKx1} 2 = \$66.\par
\masktok{MASKx29} boots\masktok{MASKx1} \$99 + \$5 = \colorbox{green!15}{\$104}.\par
\#\#\#\# \colorbox{green!15}{104}\par
The answer is: \colorbox{green!15}{104}.\masktok{MASKx1}
}
\end{tcolorbox}
\end{tcbraster}

\medskip
\textbf{Denoising Timestep \(t = 0.00\)}

\begin{tcbraster}[raster columns=2, raster equal height=rows, raster column skip=2mm]
\begin{tcolorbox}[colback=red!3,colframe=red!35!black,boxrule=0.4pt,arc=1mm]
{\ttfamily
If one pair of heels costs \$33, then the other pair costs \$33 x 2 = \$66.\par
The total cost of the two pairs of heels is \$33 + \$66 = \$132.\par
If the two pairs of heels together cost five dollars less than the boots, then the boots cost\$132 + \$5 =  \colorbox{red!15}{\$137}.\par
\#\#\#\# \colorbox{red!15}{137}\par
The answer is: \colorbox{red!15}{137}
}
\end{tcolorbox}
\begin{tcolorbox}[colback=blue!3,colframe=blue!35!black,boxrule=0.4pt,arc=1mm]
{\ttfamily
If one pair of heels costs \$33 and the other costs twice as much, then the second pair of heels costs \$33 x 2 = \$66.\par
Together, the heels cost \$33 + \$66 = \$99.\par
The heels cost five dollars less than the boots, so the boots cost \$99 + \$5 = \colorbox{green!15}{\$104}.\par
\#\#\#\# \colorbox{green!15}{104}\par
The answer is: \colorbox{green!15}{104}.
}
\end{tcolorbox}
\end{tcbraster}

\end{tcolorbox}

\vspace{0.8em}

\FloatBarrier

\subsection{Practical Overhead}
\label{sec:practical-overhead}

MetaState trains only the external recurrent interface while keeping the dLLM backbone frozen. Table~\ref{tab:rebuttal-training-time} reports the resulting offline training wall-clock on the 50k Tulu-3 subset. Tables~\ref{tab:rebuttal-throughput-latency} and \ref{tab:rebuttal-memory-overhead} report inference throughput, latency, and memory under the main GSM8K decoding setup. The current implementation increases per-sample latency by about 14--15\% and reduces raw throughput by about 15--18\%. The bf16 module footprint is about 90 MiB, and peak allocated memory rises by less than 1\% on both backbones.

\begin{table}[ht]
\centering
\small
\vspace{-1em}
\caption{Training wall-clock for the 50k Tulu-3 subset. The backbone remains frozen; the listed time is the offline cost of training the external MetaState interface.}
\label{tab:rebuttal-training-time}
\begin{tabular}{@{}l c c@{}}
\toprule
Family & Data & Runtime \\
\midrule
Dream-Instruct & 50k & 2.78 h (10012.8 s) \\
LLaDA-Instruct & 50k & 1.89 h (6787.9 s) \\
LLaDA-1.5 & 50k & 1.87 h (6742.1 s) \\
\bottomrule
\end{tabular}
\vspace{-1em}
\end{table}

\begin{table}[ht]
\centering
\small
\vspace{-1em}
\caption{GSM8K inference throughput and latency. The current implementation adds about 14--15\% latency and 15--18\% raw throughput overhead under the main decoding setup.}
\label{tab:rebuttal-throughput-latency}
\begin{tabular}{@{}l c c c c@{}}
\toprule
Family & Raw tok/s & Drop & Sample latency & Overhead \\
\midrule
Dream-Instruct & 162.66 $\rightarrow$ 134.03 & 17.60\% & 3096.82 $\rightarrow$ 3664.79 ms & +15.49\% \\
LLaDA-Instruct & 150.20 $\rightarrow$ 127.47 & 15.13\% & 3815.96 $\rightarrow$ 4427.21 ms & +13.82\% \\
\bottomrule
\end{tabular}
\vspace{-1em}
\end{table}

\begin{table*}
\centering
\small
\caption{Memory overhead. The bf16 module footprint is about 90 MiB, and peak allocated memory rises by less than 1\% for both backbones.}
\label{tab:rebuttal-memory-overhead}
\begin{tabular}{@{}l c c c c c@{}}
\toprule
Family & Module size & State / seq & Allocated memory & Delta & Peak delta \\
\midrule
Dream & 89.18 MiB & 0.125 MiB & 14526.15 $\rightarrow$ 14635.34 MiB & +109.19 MiB  & +181.50 MiB \\
LLaDA & 91.43 MiB & 0.125 MiB & 15292.51 $\rightarrow$ 15402.32 MiB & +109.81 MiB  & +117.84 MiB  \\
\bottomrule
\end{tabular}
\end{table*}

\subsection{Result on an RL-Optimized Backbone}
\label{sec:llada15-results}

To test whether MetaState transfers beyond the original LLaDA and Dream checkpoints, we attach the same frozen-backbone recurrent interface to LLaDA 1.5, a newer RL-optimized dLLM backbone. Table~\ref{tab:rebuttal-llada15} shows positive gains on all four metrics, with the largest gain on MBPP. This result is important because it shows that MetaState remains useful on a stronger post-trained dLLM, supporting the view that the method is an interface-level augmentation rather than a checkpoint-specific artifact.

\begin{table}
\centering
\small
\caption{MetaState on LLaDA 1.5, a newer RL-optimized dLLM backbone. MetaState improves all four metrics under the same protocol, supporting transfer beyond the original LLaDA and Dream checkpoints.}
\label{tab:rebuttal-llada15}
\begin{tabular}{@{}l l c c c c@{}}
\toprule
Benchmark & Original & MetaState & Gain \\
\midrule
GSM8K  & 80.74 & 82.56 & +1.82 \\
HumanEval &  37.80 & 38.41 & +0.61 \\
MBPP &  28.20 & 35.60 & +7.40 \\
MATH-500 & 37.40 & 38.00 & +0.60 \\
\bottomrule
\end{tabular}
\end{table}

\FloatBarrier

\subsection{Limitations}
\label{sec:limitations}

We discuss several limitations of the current approach.
\textbf{Training overhead.}
The $K$-step unrolling training pipeline requires $K{+}1$ sequential forward passes through the backbone per training iteration (one warmup pass plus $K$ unrolled steps), compared to a single forward pass in standard dLLM training. This multiplicative increase in computation directly raises wall-clock training time and GPU memory consumption, as intermediate activations must be retained across steps for backpropagation through the unrolled computation graph, although the recurrent modules themselves are lightweight (${\sim}0.6\%$ of backbone parameters).
\textbf{Inference overhead.}
At inference time, each denoising step requires executing the Mixer, Updater, and Injector in addition to the frozen backbone forward pass. Although these modules operate in compact bottleneck dimensions and add modest per-step latency, the overhead accumulates over the full denoising trajectory. Furthermore, maintaining the constant-size persistent state tensor increases peak memory usage during generation.
\textbf{Potential solutions.}
Several directions may mitigate the above limitations. Systems-level optimizations such as kernel fusion of the recurrent modules, hardware-aware scheduling to overlap recurrent and backbone computation, and selective recomputation strategies could reduce both training and inference overhead. Curriculum-based $K$ scheduling or auxiliary objectives that explicitly encourage long-horizon state stability may also help close the training-to-inference extrapolation gap.

\end{document}